\documentclass[final]{cvpr}
\usepackage{helvet}  
\usepackage{courier}  
\usepackage[hyphens]{url}  
\usepackage{graphicx} 
\usepackage{epsfig}
\usepackage{amsmath}
\usepackage{amssymb}
\usepackage{multirow}
\usepackage{fixltx2e}
\usepackage{amsthm}
\usepackage{color}
\usepackage{arydshln }
\usepackage{booktabs}
\usepackage{threeparttable}

\newcommand*{\modify}[1]{\textcolor{black}{#1}}

\newcommand{\myie}[1]{{\it i.e.,} #1}
\newcommand{\myeg}[1]{{\it e.g.,} #1}

\newcommand{\netcls}{AE\textsuperscript{2}INetCls}
\newcommand{\netseg}{AE\textsuperscript{2}INetSeg}
\newcommand{\module}{AE\textsuperscript{2}IL}
\newcommand{\smodule}{SymAE\textsuperscript{2}IL}
\definecolor{window}{RGB}{119,77,0}
\definecolor{wall}{RGB}{213,255,0}
\definecolor{board}{RGB}{187,136,0}
\definecolor{clutter}{RGB}{0,0,255}
\usepackage[caption=false,font=footnotesize,labelfont=sf,textfont=sf]{subfig}
\usepackage[pagebackref=true,breaklinks=true,colorlinks,bookmarks=false]{hyperref}

\begin{document}

%
\title{Adaptive Edge-to-Edge Interaction Learning for Point Cloud Analysis}
%
%
%
%

\author{Shanshan Zhao\thanks{sshan.zhao00@gmail.com},
        Mingming Gong,
        Xi Li,
        Dacheng Tao
}

%
%

\maketitle
\begin{abstract}
Recent years have witnessed the great success of deep learning on various point cloud analysis tasks, \emph{e.g.,} classification and semantic segmentation. Since point cloud data is sparse and irregularly distributed, one key issue for point cloud data processing is extracting useful information from local regions.
To achieve this, previous works mainly extract the points' features from local regions by learning the relation between each pair of adjacent points. 
However, these works ignore the relation between edges in local regions, which encodes the local shape information.
Associating the neighbouring edges could potentially make the point-to-point relation more aware of the local structure and more robust. 
To explore the role of the relation between edges, this paper proposes a novel \textbf{A}daptive \textbf{E}dge-to-\textbf{E}dge \textbf{I}nteraction \textbf{L}earning module (\module), which aims to enhance the point-to-point relation through modelling the edge-to-edge interaction in the local region adaptively. 
We further extend the \module\ module to a symmetric version, named \smodule, to capture the local structure more thoroughly. Taking advantage of the proposed modules, we develop two networks, \netseg\ and \netcls, for segmentation and shape classification tasks, respectively. Various experiments on several public point cloud datasets demonstrate the effectiveness of our method for point cloud analysis.

\end{abstract}


%
%
%
%
\section{Introduction}
\label{sec:intro}
In recent years, a lot of effort has been made to exploit deep learning to study 3D point cloud analysis tasks\cite{9740525,9735342,9440696,9661313,9018080}, which are important for real-world applications, such as autonomous driving~\cite{li2019net} and robotics manipulation~\cite{kim2014semantic}. 
Since point cloud data does not have a regular structure like images, it cannot be processed straightforwardly in the deep convolutional neural networks (DCNNs), which have been extensively and successfully applied for 2D image analysis~\cite{9356353,9320524,8821313}. 
As a result, many methods have been proposed to develop efficient and novel operations for point cloud data processing. One strategy is first voxelizing the points and obtaining the voxel-based representation, which can be processed by the conventional convolutional operations~\cite{maturana2015voxnet,wu20153d} or improved convolutions~\cite{graham20183d,choy20194d}.
However, these voxel-based approaches often suffer from  quantization
loss of the structure due to the low resolution caused by voxelization.

\begin{figure}[t]
\begin{center}
 \includegraphics[width=0.9\linewidth]{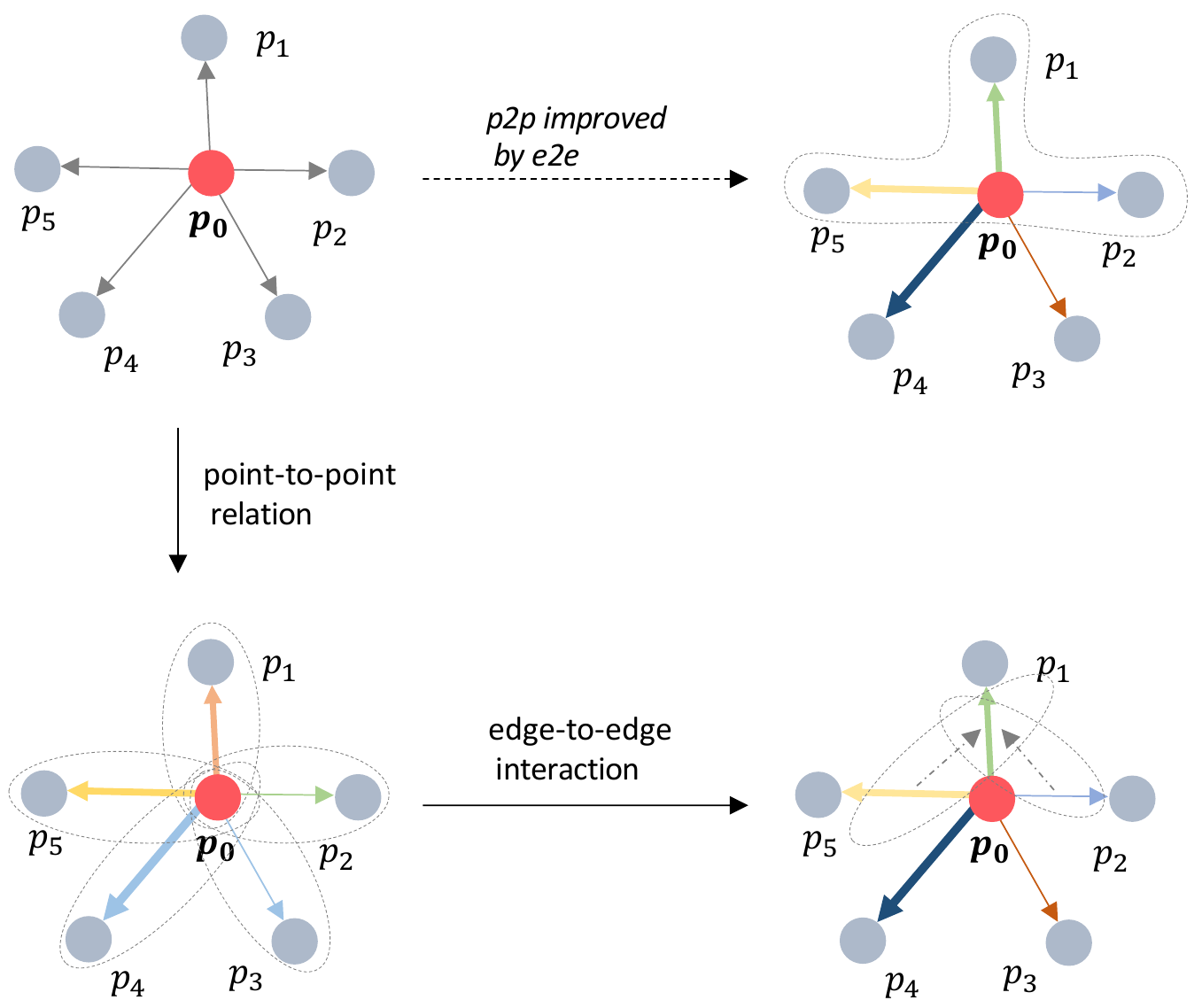}
\end{center}
   \caption{A brief illustration of our method. Previous related methods calculate the point-to-point relation with only considering the corresponding two points (left-bottom), while our proposed edge-to-edge interaction improves the point-to-point relation by further exploring the relation between edges (\myeg{$p_0\rightarrow{p_1}$}, right-bottom). As a result, we could consider that in comparison with the previous point-to-point relation learning, our method would model the points' relationship by exploiting more local information (\myeg{$p_0\rightarrow{p_1}$}, right-up).}
\label{fig:compare}
\end{figure}
Another strategy is designing a deep model that can
 learn representations from the raw point cloud directly, which this paper is focused on. The core idea of such strategy is developing an efficient operation that can process point cloud data with point-based representation as the basic component (such as, a layer) of a deep model. 
 The pioneering work of this clue, \myie{PointNet~\cite{qi2017pointnet}}, extracts the features from each point directly using the multi-layer perceptron (MLP). Despite being efficient, it omits the local structure, which is significant for learning discriminative representations. PointNet++~\cite{qi2017pointnet++}, as an extension of PointNet, attempts to model the local shape by introducing a hierarchical encoder-decoder structure with point sampling and feature propagation operations. Although considering the local structure, PointNet++ extracts the local representation from the points in the local region using the Max-Pooling function, which ignores the relation between adjacent points.
 
 Following these two works, a lot of variants~\cite{wang2019dynamic,liu2019relation,ptransformer,wang2019graph,ran2021learning} have been proposed to extract discriminative features from the local regions by developing more powerful point-to-point relation learning. 
 For example, DGCNN~\cite{wang2019dynamic} proposes the EdgeConv operation that extracts the local features from the center point and the edges (\myeg{spatial relative position}) between it and its neighbours. Another interesting work, RS-CNN~\cite{liu2019relation} maps the predefined geometric priors between two adjacent points into a high-level relation expression, and then considers it as weights to aggregate the local contextual information. In comparison with RS-CNN, which only exploits the geometric (low-level) relation, RPNet~\cite{ran2021learning} also models the semantic (high-level) relation.

\modify{As these point-based methods solely model the point-to-point relation for each pair of adjacent points, the learned representation for the edge might lack the local structure information, making the relation not discriminative and not robust. 
To improve the point-to-point relation by considering the local shape information, PointWeb~\cite{zhao2019pointweb} constructs a fully connected graph to associate all adjacent point pairs in the local region to make the point's representation share-aware. However, this method still focuses on the relation between each point pairs, which is not flexible to enhance the point-to-point relation. Since our goal is to improve the point-to-point relation, then a feasible solution is to associate the relations between different point pairs in the local region.}

\modify{To this end, we propose an Adaptive Edge-to-Edge Interaction Learning (\module) module. Specifically, for a point, we first find its $K$ neighbours, and thus there are $K$ edges emanating from it to its neighbours. Then, for each edge we consider other edges' information through modelling the edge-to-edge interaction in three steps: 1) find its nearest neighbours from the $K$ edges; 2) learn the relation between it and its neighbours; 3) use these learned relations to update its information. 
Furthermore, to model the local structure thoroughly and explore the reverse edges, we extend the \module\ to a symmetric version, namely \smodule, the details of which can be found in the third section. As illustrated in Figure~\ref{fig:compare}, in comparison with previous point-to-point relation learning, our edge-to-edge interaction is able to improve the learned points' relation by considering other learned relations in the local region.}

Taking advantage of the (symmetric) adaptive edge-to-edge interaction learning modules, we develop two networks, \myie{\netcls\ and \netseg}, for shape classification and segmentation, respectively. 
Our contributions can be summarized as follows:
\begin{itemize}
    \item We propose an adaptive edge-to-edge interaction learning module, \myie{\module}, which aims at enhancing the learned point-to-point relation, {\it i.e.,} the representation for the edge, and makes it more aware of the local structure.
    \item We extend the \module\ to a symmetric version, namely \smodule, for better capturing the local information. Then, we exploit the proposed modules to design models for point cloud classification and segmentation.
    \item The designed models outperform most of previous point-based approaches on several public point cloud datasets. Additionally, improvements could be observed by combining our strategy with previous relevant operations. All these result can demonstrate the effectiveness of our method. 
\end{itemize}


\section{Related Works}
Currently, many efforts have been made to exploit deep learning on point cloud processing. Some focus on developing efficient operations which can be applied on the raw point cloud (point-based) or voxelized point cloud (voxel-based), while some attempt to integrate the point-based solution with the voxel-based~\cite{tang2020searching}. In addition, there are also some methods that mainly focus on the 3D segmentation task by proposing novel voxelization strategy~\cite{zhu2021cylindrical}, combining the voxel and mesh representations~\cite{hu2021vmnet}, designing efficient multi-modal fusion strategy~\cite{hu2021bpnet}, introducing the stratified strategy to capture long-range contexts~\cite{Lai_2022_CVPR}, studying the representations in boundaries~\cite{Tang_2022_CVPR}, or
exploring the pyramid architecture for processing multi-scale representations~\cite{Nie_2022_CVPR}. These methods achieve remarkable performance for point cloud segmentation. In this paper, we aim to develop an efficient operation by considering the relationships between edges as well as the relations between points, and then build the point cloud analysis networks upon the designed basic operation. Therefore, in this section, we mainly review those point-based methods which are also focused on the basic operation designing. We refer to the survey~\cite{guo2020deep} for a thorough understanding of point cloud analysis.

Following PointNet~\cite{qi2017pointnet}, which is the first attempt to apply deep learning directly on the sparse and unstructured point sets, and its extension PointNet++~\cite{qi2017pointnet++}, a lot of efforts~\cite{landrieu2018large,wu2019pointconv,lang2020samplenet,Chen_2022_CVPR,liu2019dynamic,liu2020closerlook3d,wang2018deep,wongefficient,nie2021differentiable,Le_2020_CVPR,Nezhadarya_2020_CVPR,wang2019exploiting,hua2018pointwise,zhang2021BowPooling,9051667,liu2019point2sequence,xu2018spidercnn,komarichev2019cnn,trans3d} have investigated the feature extraction of the local structure. Most of these point based works mainly focus on one or more of the following components: 1) points sampling, 2) relation learning, and 3) convolutional operation.

\noindent 
{\textbf{Points Sampling}}. To capture the contextual information in a hierarchical structure, PointNet++~\cite{qi2017pointnet++} exploits the {\it farthest point sampling} (FPS) algorithm to sample a subset from the input points. Since the FPS algorithm is permutation-variant and samples points from low-dimension Euclidean space, PAT~\cite{yang2019modeling} proposes Gumbel Subset Sampling to select the subset, which is more robust to outliers. In comparison, to overcome the issue existing in FPS, PointASNL~\cite{yan2020pointasnl} proposes an adaptive sampling strategy to refine the initial sampled points, which considers both low- and high-dimension embedding space. Interestingly, a recent work, RandLA-Net~\cite{hu2020randla} compares several point sampling approaches, and observes that the Random Sampling strategy is more suitable for large-scale point clouds.

\noindent 
\textbf{Relation Learning}. To represent the local structure, PointNet++~\cite{qi2017pointnet++} extracts the features of each point in the local region using the MLP and then exploits the Max-Pooling operation to get the local region feature vector. However, it ignores the geometric relationships between points, which causes the limitation on the modeling of local structures. To improve this, DGCNN~\cite{wang2019dynamic} exploits the proposed EdgeConv on the constructed local neighbourhood graph to model local geometric structures. EdgeConv aggregates the features of the edges emanating from the central point of the local region as its new representation. In comparison, some works aim to map the edge features into weights for feature association. For example, extending regular 2D CNN to irregular configuration, RS-CNN~\cite{liu2019relation} encodes the predefined geometric priors, \myeg{the spatial distance,} between two adjacent points as a high-level relation expression, \myie{weight vector}. Similarly, GAC~\cite{wang2019graph} learns attentional weights from both spatial and feature distances. One point in common among these works is that in a local region, only the edges connecting the central point to the others are considered. In comparison, PointWeb~\cite{zhao2019pointweb} constructs a densely-connected graph and aims to find the interaction between all adjacent points for better description of the local structure. PointTransformer~\cite{ptransformer} is built on the remarkable transformer technique~\cite{vaswani2017attention} and achieves the state-of-the-art indoor segmentation performance on the S3DIS dataset. 
While all these papers focus on the learning of point-to-point relation, this paper attempts to learn the edge-to-edge interaction adaptively for the enhancement of the point-to-point relation.

\noindent 
{\textbf{Convolutional Operation.}} Motivated by the 2D convolution kernel, KPConv~\cite{thomas2019kpconv} designs a set of 3D kernel points. The kernel points are used to define the area where the kernel weights are applied to extract the features. A recent work, PAConv~\cite{paconv}, proposes to construct the convolutional kernels by dynamically assembling basic weight matrices in a pre-defined Weight Bank.
In addition, some works attempt to apply projection in the feature space so that the projected features can be processed by the standard convolutional operation directly. Fox example, PointCNN~\cite{li2018pointcnn} transforms the points in a local region to the canonical order through learning a transformation matrix and then the traditional convolution can be applied. In comparison with PointCNN, FPConv~\cite{lin2020fpconv} learns a weight map to softly project local points onto a 2D grid, which is further processed by the regular 2D convolutional operation.


\section{Our Method}
\begin{figure*}
\begin{center}
\includegraphics[width=0.9\linewidth,height=3.0in]{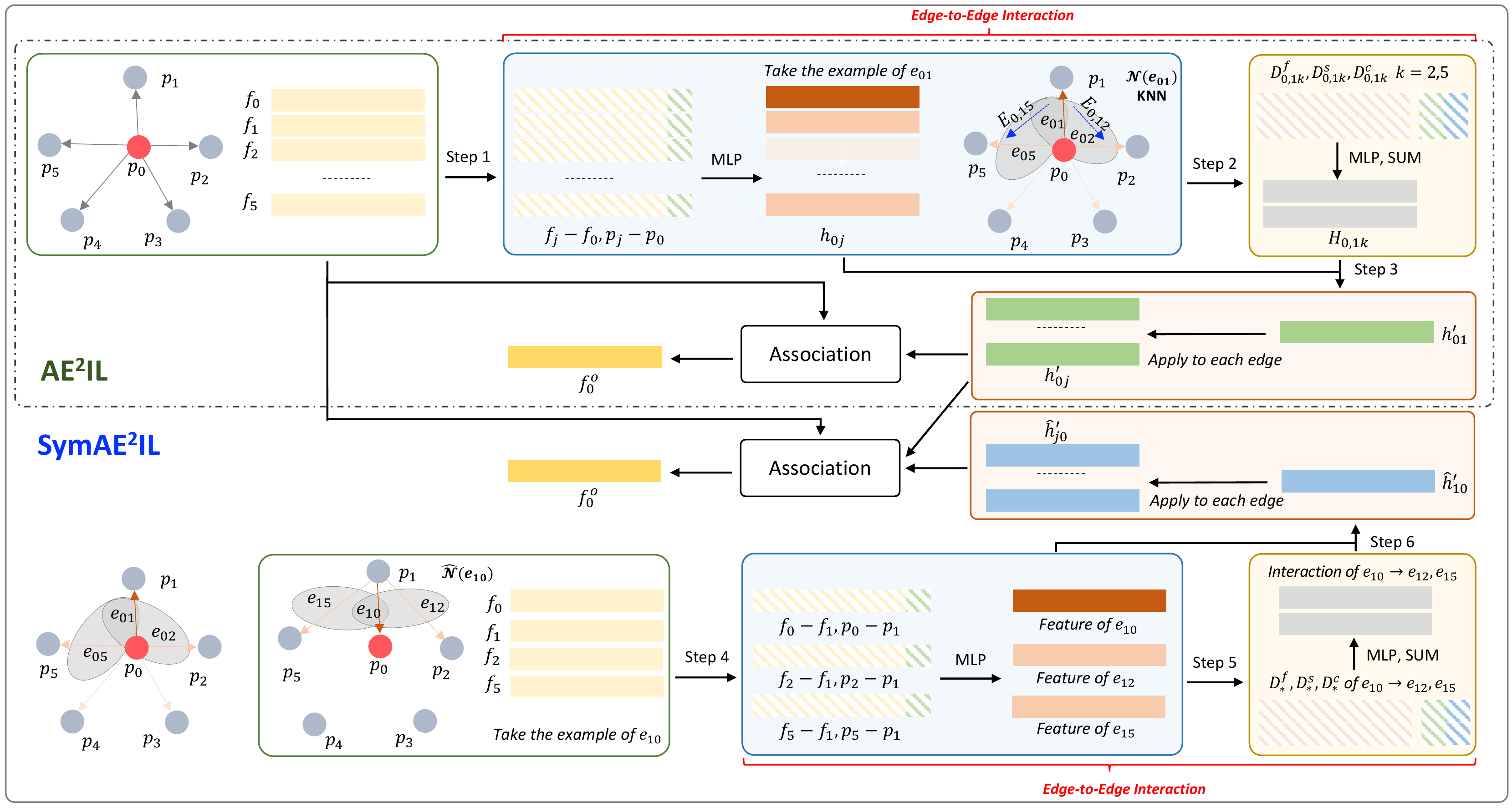}
\end{center}
   \caption{Illustration of \module\ and \smodule. \module\ marked by the gray dotted line consists of Step 1, Step 2, and Step 3, while \smodule\ marked by the gray solid line contains all steps. The edges located in one gray ellipse are neighbours. In this example, $p_0$ is the central point, and $p_1$-$p_5$ are its neighbours. Note that, from Step 2 to Step 6, we take the example of the edges $e_{01}$ and $e_{10}$. The blue dotted arrow denotes the $edge$ between edges. The two association operations are formulated as Eq.~\ref{eq:murho} (for the top one) and Eq.~\ref{eq:t2} (for the below one), respectively. Notations are identical to the text. Best viewed in color (zoom in for details).}
\vspace{-3mm}
\label{fig:module}
\end{figure*}

In this section, we present the proposed novel adaptive edge-to-edge interaction learning modules, \module\ and its symmetric version \smodule\ in detail. In addition, we also provide an analysis for the differences between an existing work and our approach in the last.


\subsection{\module\ Module}
\label{sec:dril}

Let $\mathcal{P}=\{p_1,p_2,...,p_N\}\subset \mathbb{R}^{3}$, where $p_i$ represents the point's spatial position, denote the processed point cloud consisting of $N$ points, $\mathcal{F}=\{f_1,f_2,...,f_N\}\subset \mathbb{R}^{C_{in}}$ denote the corresponding feature set, where $C_{in}$ is the number of channels. Our new \module~ module takes $\mathcal{P}$ and $\mathcal{F}$ as the input, and outputs the enhanced representation $f^o_i\in \mathbb{R}^{C_{out}}$ for each point $p_i$ in $\mathcal{P}^s$. $\mathcal{P}^s$ is a subset sampled from $\mathcal{P}$ via the FPS technique~\cite{qi2017pointnet++}, and $C_{out}$ is the channel number.


Specifically, for a certain point $p_i$, we find its $K$ neighbours $\mathcal{N}(p_i)\subseteq \mathcal{P}$ from $\mathcal{P}$. This is implemented by the K-nearest neighbour (K-NN) algorithm according to the spatial distance. To extract the local features, we follow \cite{liu2019relation,wang2019exploiting,wang2019graph} to first encode the point-to-point relation ($h_{ij}$) based on the spatial relative  position and difference between features as:
\begin{equation}
\begin{aligned}
    h_{ij} 
        = \sigma([(p_{j}-p_i) || (f_{j}-f_i])),\ p_{j}\in \mathcal{N}(p_i),
\end{aligned}
\label{eq:sigma}
\end{equation}
where $[\cdot||\cdot]$ is the concatenation operation, and $\sigma(\cdot)$ denotes a mapping function. In the paper, we use MLP as the function. Given the learned point-to-point relation $h_{ij}$, we can integrate the point feature $f_i$ and its $K$ neighbouring features $f_j$ following ~\cite{liu2019relation,wang2019exploiting,wang2019graph}. However, as discussed before, only studying the relations between two points solely may fail to model the local structure well. We thus seek a solution by adaptively investigating the interaction between the edges.

In detail, for the central point $p_i$ we have $K$ directed edges\footnote{Note that, in our paper the edge $e_{ij}$ between $p_i$ and $p_j$ is directed, where $p_i$ is the staring / emanating point and $p_j$ is the terminal point.}. Among them, the edge $e_{ij}$ emanates from point $p_i$ to its neighbour $p_j\in \mathcal{N}(p_i)$. We then utilize K-NN algorithm to find the $K_e$ nearest neighbours for each edge according to the distance between edges. Here, the distance between $e_{ij}$ and $e_{ik}$ is computed as:
\begin{equation}
    Dist(e_{ij}, e_{ik}) = \sqrt{||p_j - p_k||_2},
\end{equation}
{\it i.e.,} the spatial Euclidean distance between the terminal points
of the edges.
For a specific edge $e_{ij}$, 
we denote its neighbours as $e_{ik}\in \mathcal{N}(e_{ij})$, where $e_{ik}$ represents the edge from $p_i$ to its neighbour $p_k\in \mathcal{N}(p_i)$. We take $E_{i,jk}$ to represent the \emph{edge} from $e_{ij}$ to $e_{ik}$, and define $H_{i,jk}$ as their interaction. These symbolic marks are illustrated in Figure~\ref{fig:module}.



To obtain $H_{i,jk}$, we extract the information of \emph{edge} $E_{i,jk}$. Here, we consider three kinds of information. The first two are the spatial relative position $D^s_{i,jk}$ and difference between features $D^f_{i,jk}$. Before computing $D^f_{i,jk}$, we first use MLPs ($\phi$ and $\psi$) to encode the edges' features,
and then calculate the two relations as:
\begin{equation}
    \begin{aligned}
        & D^s_{i,jk} = p_k - p_j, \\ 
        & D^f_{i,jk} = \phi(h_{ik}) - \psi(h_{ij}).
    \end{aligned}
\label{eq:phipsi}
\end{equation}

 Apart from $D^s_{i,jk}$ and $D^f_{i,jk}$, we further compute the surface normal $D^c_{i,jk}$ as follows:
\begin{equation}
    D^c_{i,jk} = (p_k-p_i) \times (p_j - p_i),
\label{eq:normal}
\end{equation}
where $\times$ denotes the cross product operation. 
By taking $D^s_{i,jk}$, $D^f_{i,jk}$, and $D^c_{i,jk}$ as the inputs, we first encode the spatial information into a new feature vector and then capture $H_{i,jk}$ via a summation operation:
\begin{equation}
    H_{i,jk} = D^f_{i,jk} + \gamma(D^s_{i,jk}||D^c_{i,jk}),
\label{eq:gamma}
\end{equation}
where $\gamma$ is an MLP.
Therefore, the edge-to-edge interaction $H_{i,jk}$ not only encodes the relations between edges in both low- (Euclidean) and high- (feature) dimensional space, but also considers the plane information. We refer to the ablation studies for more analysis.


The next goal is to update the point-to-point relation $h_{ij}$ using the learned edge-to-edge interactions between $e_{ij}$ and its $K_e$ neighbours $\mathcal{N}(e_{ij})$ via the attention technique~\cite{hu2020randla,vaswani2017attention}. In specific, we first calculate the attentional weights as follows:
\begin{equation}
\begin{aligned}
    & w_{i,jk} = \alpha(H_{i,jk}),\\
     & w^{'}_{i,jk} =  \frac{exp(w_{i,jk})}{\sum_{l\in \dot{\mathcal{N}}(e_{ij})}exp(w_{i,jl})}, \\
\end{aligned}
\label{eq:alpha}
\end{equation}
where $\alpha$ is an MLP, and $\dot{\mathcal{N}}(e_{ij})$ denotes an index set containing the index of the terminal point of the edges in $\mathcal{N}(e_{ij})$. We then aggregate $\{h_{ik}\}_{k\in \dot{\mathcal{N}}(e_{ij})}$ and the spatial relation $\gamma(D^s_{i,jk}||D^c_{i,jk})$ using the learned attentional weights as follows:
\begin{equation}
    \begin{aligned}
        & h^{'}_{ij} = \sum_{k\in \dot{\mathcal{N}}(e_{ij})} w^{'}_{i,jk} \cdot (\beta(h_{ik})+\gamma(D^s_{i,jk}||D^c_{i,jk})), 
    \end{aligned}
\label{eq:beta}
\end{equation}
where $\cdot$ denotes the element-wise multiplication operation, and $\beta$ is an MLP. 

Now, we achieve the interaction between the edges emanating from $p_i$, and get the enhanced representation $h_{ij}^{'}$ for each directed edge $e_{ij}$. Lastly, we employ three consecutive operations, \myie{a shared MLP $\mu$ and a max-pooling operation}, and a residual connection~\cite{hu2020randla,he2016deep}, to update the features of $p_i$ so that it contains the extracted local structure information, \myie{}
\begin{equation}
    \begin{aligned}
    &f_{ij} = \mu([f_i||h_{ij}^{'}]), \\
    &\langle f^{o}_i \rangle_c = \max_{j=1,2,...,K} \langle f_{ij} \rangle_c + \langle \rho(f_i) \rangle_c, 
    \end{aligned}
\label{eq:murho}
\end{equation}
where $c\in[1,2,...,C_{out}]$, $\rho$ is an MLP, and $\langle f \rangle_i$ is the $i^{th}$ element of the feature vector $f$.

\subsection{\smodule\ Module}
\label{sec:symdril}
For a local region centering on point $p_i$, \module\ exploits the local structure to enhance the point-to-point relation $h_{ij}$ through learning the edge-to-edge interactions between $e_{ij}$ and its neighbours $\mathcal{N}(e_{ij})$. Taking an example of one of its neighbours $e_{ik}$ ($j\neq k$) starting from $p_i$ to $p_{k}$. \module\ models the interaction between $e_{ij}$ and $e_{ik}$, but overlooks the reverse edge $e_{ji}$ and the edge $e_{jk}$. We find that further studying the interaction between $e_{ji}$ and $e_{jk}$ could exploit the local structure better. From another perspective, through modeling the two interactions, we can also extract the structure information contained in the triangle constructed by $p_i$, $p_j$, and $p_k$.
We call this new module as \smodule, \myie{Symmetric Adaptive Edge-to-Edge Interaction Learning}, which simultaneously improves the representations of  $e_{ij}$ and $e_{ji}$. 

To achieve this, we first define a new edge set $\hat{\mathcal{N}}(e_{ji})$, which contains all edges emanating from $p_j$ to the terminal point of the edges in $\mathcal{N}(e_{ij})$, as the neighbours of the edge $e_{ji}$. Then,  we compute the feature of each edge in $\hat{\mathcal{N}}(e_{ji})$, and update the feature of $e_{ji}$ as we do
to update the feature of $e_{ij}$, \myie{from Eq.~\ref{eq:phipsi} to Eq.~\ref{eq:beta}}. This process is illustrated by the boxes in the bottom of Figure~\ref{fig:module}. Denoting the output feature of $e_{ji}$ by $\hat{h}_{ji}^{'}$, then we reformulate Eq.~\ref{eq:murho} as:
\begin{equation}
\begin{aligned}
&f_{ij} = \mu([f_i||(h_{ij}^{'}+\hat{h}_{ji}^{'})]),\\
&\langle f^{o}_i \rangle_c = \max_{j=1,2,...,K} \langle f_{ij} \rangle_c + \langle \rho(f_i) \rangle_c.
\end{aligned}
\label{eq:t2}
\end{equation}

We can use \module\ or \smodule\ as a basic operator to construct deep networks for point cloud analysis, including shape classification and segmentation. In our experiments, we use \smodule\ as the core operation, and we will study the two modules in the ablation. 
The networks for segmentation and classification are named as \netseg\ and \netcls, respectively.  The structures of \netcls\ and \netseg\ are shown in Figure~\ref{a:fig:net}.
The detailed architectures of \module, \smodule, \netcls, and \netseg\ are provided in the Supplementary Material.
\begin{figure}[t]
\begin{center}
 \includegraphics[width=0.9\linewidth,height=1.5in]{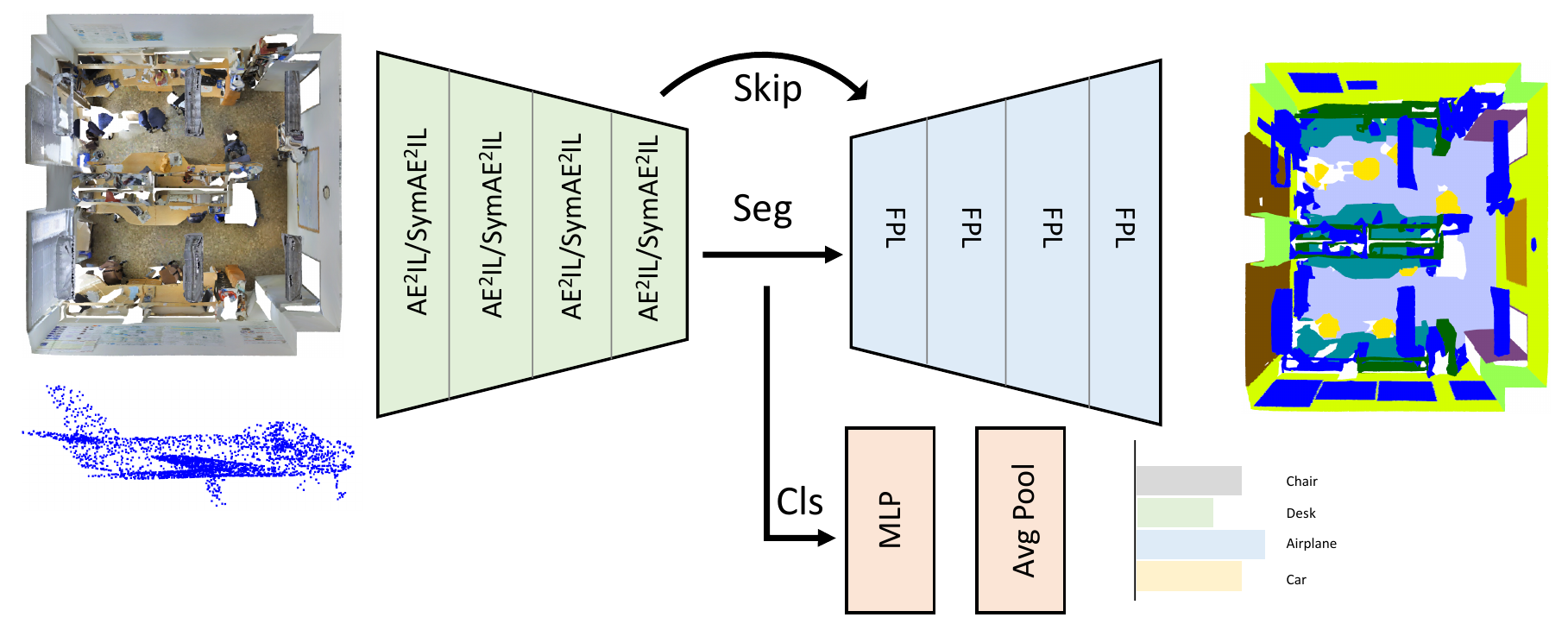}
\end{center}
   \caption{\netcls\ and \netseg. FPL, Seg, Cls, and Skip denote feature propagation layer~\cite{qi2017pointnet++}, segmentation, classification, and skip connections, respectively.}
\label{a:fig:net}
\end{figure}

\subsection{Relation to PointWeb}
In our paper, we consider the edge-to-edge interaction, which involves the point-to-point relation between the neighbours of the central point as well as the central point and its neighbours. Technically, the feature of an edge are updated with its neighbouring edges, while not in most of existing works. However, an interesting work, PointWeb~\cite{zhao2019pointweb}, 
develops the Adaptive Feature Adjustment (AFA) module to update all points in the local region before the feature aggregation. As a result, the point-to-point relation in PointWeb can be viewed as being updated. Here, we analyze the differences between the AFA strategy in PointWeb and ours.

Firstly, we consider a simplified implementation of our \module, defined as:
\begin{equation}
    \begin{aligned}
        & h^{'}_{ij} = \sum_{k\in \dot{\mathcal{N}}(e_{ij})} \omega(h_{ik}-h_{ij}) \cdot h_{ik}, 
    \end{aligned}
\label{eq:simversion}
\end{equation}
where $h_{ij}=\sigma(f_{j}-f_i)$, and $\omega$ is an MLP followed by a \textit{SoftMax} function. 
We can re-write the equation as:
\begin{flalign}
    h^{'}_{ij}   
                 = \sum_{k\in \dot{\mathcal{N}}(e_{ij})} \delta(\tau(f_{k} - f_{j})) \cdot \sigma(f_{k}-f_i),
\label{eq:de2i}
\end{flalign}
where $\delta$ denotes the \textit{SoftMax} function and $\tau$ is an MLP. The derivation is provided in the Supplementary Material. Similarly, the symmetric version of the simplified \module\ can be written as:
\begin{flalign}
    h^{'}_{ij}   = & \sum_{k\in \dot{\mathcal{N}}(e_{ij})} \delta(\tau(f_{k} - f_{j})) \cdot \sigma(f_{k}-f_i) + \nonumber\\
                 & \sum_{k\in \dot{\mathcal{N}}(e_{ij})} \delta(\tau'(f_{k} - f_i)) \cdot \sigma'(f_{k}-f_{j}),
\label{eq:sde2i}
\end{flalign}
where $\tau'$ and $\sigma'$ are both MLPs. Using our notations, we can represent the $h^{'}_{ij}$ in AFA module as follows:
\begin{flalign}
    h^{'}_{ij}  = &f^{'}_{j} - f^{'}_i \nonumber\\
                = &\sum_{k\in \dot{\mathcal{N}}(e_{ij})} [\xi(f_{j})(f_{j}-f_{k})+\xi(f_i)(f_{k}-f_{i}) +\nonumber\\
                &\xi(f_{k})(f_i-f_{j})] + h_{ij},
\label{eq:pointweb}
\end{flalign}
where $\xi$ is an MLP. The derivation is provided in the Supplementary Material.

Taking an example of three points $\{p_i, p_{j}, p_{k}\}$, where $p_j,p_k\in \mathcal{N}(p_i)$ and $k\in \dot{\mathcal{N}}(e_{ij})$, we analyze the differences between PointWeb and ours. Firstly, comparing Eq.~\ref{eq:de2i}, Eq.~\ref{eq:sde2i}, and Eq.~\ref{eq:pointweb}, we can find that when using the edge $e_{ik}$ to update $e_{ij}$, our method exploits the relation between the two edges, which is the difference between the terminal points' features in the simplified version. Although PointWeb also exploits the neighbouring edges, it does not involve the interaction between the edges. Instead, it sums the edges' features directly taking the points' features as the weights, or Eq.~\ref{eq:pointweb} can be explained as that it sums the new points' features taking the difference between points' features as the weights. In comparison, we update the edge's feature according to the relation between edges adaptively. In addition,  through modelling the edge-to-edge interaction directly, we can exploit complicated functions (Eq.~\ref{eq:phipsi}-Eq.~\ref{eq:beta} \textit{v.s.} Eq.~\ref{eq:simversion}), such as the learning of both low- and high-level relations between edges, easily. 
As a result, in comparison to PointWeb, our method is able to learn the edge-to-edge interaction more effectively and thus model the local structure better.


\section{Experiments}
\label{sec:exp}
To examine the effectiveness of our point cloud analysis approach, we conduct experiments on several tasks, including semantic segmentation, part segmentation, and classification, on widely studied benchmarks, such as  S3DIS~\cite{armeni20163d}, ScanNet v2~\cite{dai2017scannet}, SemanticKitti~\cite{semantickitti},  ShapeNetPart~\cite{Yi16}, ModelNet40~\cite{wu20153d}, and ScanObjectNN~\cite{scanobjectnn}. We provide the comparisons with previous point-based methods which also focus on developing efficient point cloud operations. We also show the effectiveness of our method by exploiting our edge-to-edge interaction to improve the point-to-point relation learning of several previous methods.
The detailed results and ablations are reported in the following.

\subsection{Implementation Details}
We train all networks with an initial learning rate of $0.1$ using the SGD optimization algorithm. For the semantic segmentation task on S3DIS (ScanNet v2, SemanticKitti), we train for $100$ ($1000$, $100$) epochs with a batch size of $8$ ($20$, $20$) and decay the learning rate by $0.1$ after $60$ ($600$, $60$) epochs and $80$ ($800$, $80$) epochs, respectively. For classification (ModelNet40 and ScanObjectNN) and part segmentation (ShapeNetPart) tasks, we reduce the learning rate until $1e-3$ using the cosine annealing~\cite{loshchilov2016sgdr} policy. We train the networks for $250$ / $250$ / $200$  epochs with a batch size of $32$ / $32$ / $16$ on ModelNet40 / ScanObjectNN / ShapeNetPart.  Following previous works~\cite{paconv,liu2019relation,qi2017pointnet++,thomas2019kpconv}, we exploit data augmentations. In specific, for classification and part segmentation, we augment the point cloud with 1 ) random anisotropic scaling in a range from $-0.66$ to $1.5$ and 2) random translation in a range from $-0.2$ to $0.2$, while for semantic segmentation, we exploit random rotation along the vertical axis, scaling in a range from $0.8$ to $1.1$, and gaussian  jittering.

 \begin{table}[!t] \footnotesize
\centering
\caption{Semantic Segmentation Performance on S3DIS  
}
\begin{threeparttable}
\setlength{\tabcolsep}{1.6mm}{
\begin{tabular}{ccccccc}
\toprule
\multirow{2}{*}{Method} & \multicolumn{3}{c}{6-fold} & \multicolumn{3}{c}{Area 5} \\ \cline{2-7}
& oA & mAcc & mIoU & oA & mAcc & mIoU \\ 
\midrule
PointNet &  78.6 &66.2 &47.6 &  - & 49.0 &41.1  \\
PointNet++ &  81.0& 67.1 &54.5&  - & - & -  \\
PointCNN & 88.1 &75.6 &65.4 &  85.9 & 63.9 & 57.3 \\
DGCNN &  84.1 &- &56.1 &  - & - & - \\
PointWeb &   87.3& 76.2& 66.7&  87.0 & 66.6 & 60.3 \\
HPEIN & 88.2 & 76.3 & 67.8&  87.2 & 68.3 & 61.9  \\
KPConv & - &79.1& 70.6 &  - & 72.8 & 67.1 \\
FPConv &  - & - & 68.7 &  88.3 & 68.9 & 62.8 \\
SegGCN & - & - & - & 88.2 & 70.4 & 63.6 \\
RandLANet&  88.0 & 82.0 &70.0&  - &- &- \\
Point2Node &  89.0 &79.1 &70.0 &88.8 &70.0 &63.0 \\
PAConv & - & 78.7 & 69.3 & - &73.0 & 66.6 \\
AdaptConv & - & - & - &90.0 & 73.2 & 67.9 \\
FPntTrans1 & - & - & - & - & 75.5& 68.5 \\
FPntTrans2 & - & - & - & - & 77.3 &70.1 \\
PntTrans (PT)  &90.2& 81.9& 73.5 & 90.8 & 76.5 & 70.4 \\
PT+\smodule & - & - & - & 90.8 & 77.4 & 70.8 \\
\midrule
\multicolumn{7}{c}{Ours} \\
\midrule
Baseline & - & - & - & 88.5 & 70.4 & 64.6 \\
+\module & - & - & - & 89.0 & 72.9 & 66.1 \\
+\smodule* & 89.9 & 82.6 & 73.7 & 89.9 & 74.3 & 68.0 \\
RSConv & - & - & - & 87.2 & 67.5 & 61.3\\
+\smodule & - & - & - & 88.5 & 69.9 & 64.1 \\
GAConv & - & - & - & 86.5 & 68.3 & 60.8 \\
+\smodule & - & - & - & 87.9 & 72.4 & 65.3\\
RandLA & - & - & - & 87.2 & 69.0 & 62.0 \\
+\smodule & - & - & - & 88.6 & 70.7 &  65.0 \\
\bottomrule
\end{tabular}}
\begin{tablenotes}
    \item \textit{+\smodule* indicates our final model \netseg; Baseline indicates the model without \module\ or \smodule.}
\end{tablenotes}

\end{threeparttable} 

\label{tab:s3dis}
\end{table}
 \begin{figure}[t]
\begin{center}
 \includegraphics[width=0.98\linewidth]{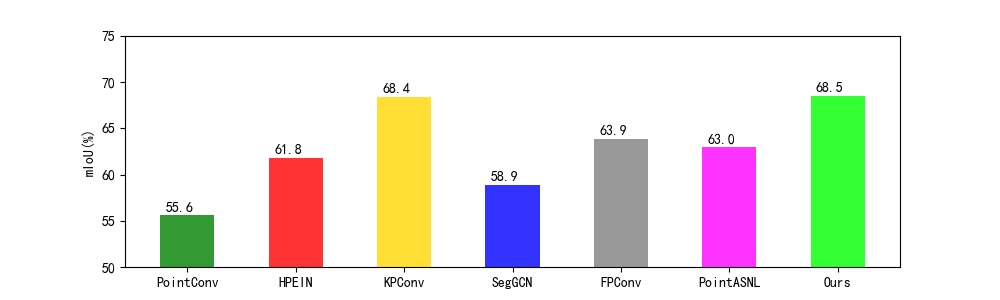}
\end{center}
   \caption{The mIoU on ScanNet v2 test set~\cite{dai2017scannet}. We make comparisons against PointConv~\cite{wu2019pointconv}, HPEIN~\cite{jiang2019hierarchical}, KPConv~\cite{thomas2019kpconv}, SegGCN~\cite{Lei_2020_CVPR}, FPConv~\cite{lin2020fpconv}, and PointASNL~\cite{yan2020pointasnl}.}
\label{fig:scannet_miou}
\end{figure}
 \begin{table} \footnotesize
\centering
\caption{Semantic Segmentation Performance on  ScanNet v2 Validation Set 
}
\begin{threeparttable} 
\setlength{\tabcolsep}{1.6mm}{
\begin{tabular}{cccc}
\toprule
{Method} &  oA & mAcc & mIoU \\ 
\midrule
Baseline & 87.3 & 77.1 & 66.0 \\
+\module & 87.5 & 77.3 & 66.5  \\
+\smodule* & 88.4 & 78.7 & 68.8\\
RSConv & 85.1 & 72.6 & 61.7\\
+\smodule & 87.4 & 77.1 & 66.5 \\
GAConv & 83.6 & 69.8 & 58.3  \\
+\smodule & 86.0 & 74.0 & 62.8\\
RandLA &  81.9 & 67.1 & 55.6 \\
+\smodule &  85.2 & 73.2 &  61.4 \\
\bottomrule
\end{tabular}
}
\begin{tablenotes}
    \item \textit{+\smodule* indicates our final model \netseg; Baseline indicates the model without \module\ or \smodule.}
\end{tablenotes}

\end{threeparttable} 

\label{tab:scannet}
\end{table}
\subsection{3D Scene Semantic Segmentation}
Here, we study 3D scene segmentation on two indoor 3D segmentation datasets, \myie{S3DIS~\cite{armeni20163d} and ScanNet v2~\cite{dai2017scannet}}, and one outdoor 3D segmentation dataset SemanticKitti~\cite{semantickitti} to evaluate the capacity of \netseg.

S3DIS contains 271 rooms captured from 6 areas. It provides 3D points and their corresponding RGB values. Each point is annotated with one of the semantic labels from 13 categories, such as table, wall, and sofa. In training time, we randomly select 14,000 points from a $2m\times 2m$ block on-the-fly. Each point is represented as a 9-dim vector with XYZ, RGB, and normalized position in the room. All points are evaluated at test time. We study two settings for the task, \myie{6-fold cross-validation and Area 5 validation.} In Table~\ref{tab:s3dis},
we make comparisons against the previous methods, including PointNet~\cite{qi2017pointnet}, PointNet++~\cite{qi2017pointnet++}, PointCNN~\cite{li2018pointcnn}, DGCNN~\cite{wang2019dynamic}, PointWeb~\cite{zhao2019pointweb}, HPEIN~\cite{jiang2019hierarchical}, KPConv~\cite{thomas2019kpconv}, FPConv~\cite{lin2020fpconv}, SegGCN~\cite{Lei_2020_CVPR}, RandLANet~\cite{hu2020randla}, Point2Node~\cite{han2020point2node}, 
PAConv~\cite{paconv}, PointTransformer (PntTrans)~\cite{ptransformer}, AdaptConv~\cite{zhou2021adaptive}, and FastPointTransformer (FPntTrans1,2)~\cite{fpt}. To further show that our edge-to-edge interaction is able to enhance the point-to-point relation, we also re-implement previous point cloud analysis operations, including RSConv~\cite{liu2019relation}, GAConv~\cite{wang2019graph}, and RandLA~\cite{hu2020randla} in our framework, and use our strategy to improve the learned relationships between adjacent points.
Following previous works, like KPConv, PAConv, and AdaptConv, we use the voting scheme during inference.
\modify{As shown in Table~\ref{tab:s3dis},
our model performs better than almost all of previous works on all metrics, including overall Accuracy (oA), mean IoU (mIoU), and mean class Accuracy (mAcc). 
From the table, we can observe that currently methods utilizing the local transformer achieve high-performing performance. Specifically, PntTrans yields highest scores on Area 5, while lower scores than ours on 6-fold. 
FPntTrans also achieves higher scores than ours. However, these two transformer-based methods both exploit the voxelization technique in order to process the whole scene,
so the performance might be impacted by the voxel size. For example, as shown in Table~\ref{tab:s3dis}, 
FPntTrans1 (voxel size: 5cm) gets lower scores than FPnTrans2 (voxel size: 4cm), which means
the increase in voxel size is likely to result in performance degradation.
In addition, 
in comparison with the top three methods, \myie{AdaptConv (15.8M), PntTrans (7.8M), and FPntTrans (37.9M)}, our model has fewer parameters (3.6M).
More details can be found in Sec.~\ref{sec:comp}. 
The improvements over previous point analysis operations can further show the capability of the proposed module. 
We also attempt to use our edge-to-edge interaction to improve the attentional weights in PointTransformer (
\myie{PT+\smodule}), and an improvement can also be observed. 
Finally, it is worth noting that our method outperforms PointWeb~\cite{zhao2019pointweb} by a large margin, which demonstrates the superiority of the proposed edge-to-edge interaction.
The scores for each class are given in the Supplementary Material.}

For ScanNet v2~\cite{dai2017scannet}, we train the model on the training set ($1201$ scans), and make evaluation on the test set ($100$ scans). 
There are $20$ meaningful categories and one class for free space.
During training, we randomly sample $14,000$ points from a $2m\times 2m$ block on-the-fly. Each point 
is represented as a 6-dim vector with XYZ and RGB. 
We mainly compare our model with previous point-based methods. As shown in Figure~\ref{fig:scannet_miou}, our model performs better than most of methods by a large margin. In the benchmark website, we can find some methods yield higher scores through exploring other clues, such as, rendering virtual views~\cite{kundu2020virtual}, training multiple tasks~\cite{hu2020jsenet}, and combining 2D and 3D domains~\cite{hu2021bpnet}. We also provide the performance of previous point cloud analysis operations with / without our edge-to-edge interaction module on the validation set of ScanNet v2. As shown in Table~\ref{tab:scannet}, our module could bring remarkable improvements.
See the detailed scores for each class in the Supplementary Material. 

\begin{figure}[t]
\begin{center}
 \includegraphics[width=0.98\linewidth]{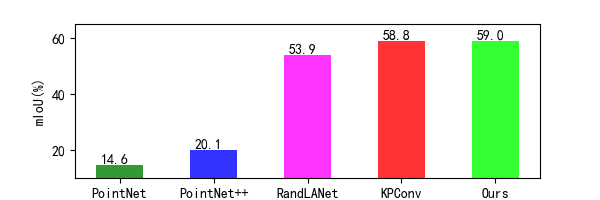}
\end{center}
   \caption{The mIoU on SemanticKitti test set~\cite{semantickitti}. We compare our method with previous point-based methods, such as PointNet~\cite{qi2017pointnet}, PointNet++~\cite{qi2017pointnet++}, RandLANet~\cite{hu2020randla}, and KPConv~\cite{thomas2019kpconv}. }
\label{fig:sk_test}
\end{figure}

\begin{figure}[!t]
\begin{center}
 \includegraphics[width=0.98\linewidth]{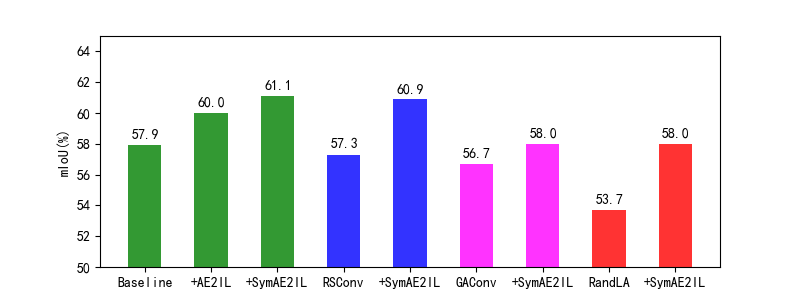}
\end{center}
   \caption{The mIoU on SemanticKitti validation set~\cite{semantickitti}. As we do on S3DIS and ScanNet v2, we study our module by adding it into the Baseline and other point cloud analysis operations. +\smodule* indicates our final model \netseg, while Baseline indicates the model without \module\ or \smodule.}
\label{fig:sk_val}
\end{figure}

For SemanticKitti~\cite{semantickitti}, which is a public large-scale LiDAR semantic segmentation dataset consisting 21 of LiDAR scans sequences with dense semantic annotations, we train the model on the sequences $00\thicksim07$ and $09\thicksim10$, and conduct validation on the sequence 08. We test the performance on sequences $11\thicksim21$ by submitting the predicted results online. There is no RGB color information and are only 3D coordinates available. During training, we randomly sample $14,000$ points from each selected frame. Following RandLANet~\cite{hu2020randla}, during testing, for a LiDAR frame, we do multiple times of randomly sampling $14,000$ points to guarantee that all points have predictions. The performance on the test set is reported in Figure~\ref{fig:sk_test}, and our method achieves highest performance among the point-based methods. The comparisons between previous point cloud analysis operations without our module and with our module in Figure~\ref{fig:sk_val} can also show the effectiveness of our method. Detailed scores for each category are provided in the Supplementary Material.
 
\subsection{3D Shape Part Segmentation}
The shape part segmentation task aims to predict part category label for each point in a 3D model. We evaluate the proposed \netseg\ on the ShapeNetPart dataset~\cite{Yi16}. There are $16,881$ CAD models from $16$ object categories, which are labeled with $50$ parts in total. Following~\cite{li2018pointcnn,zhang2019shellnet}, we split the models into $14,006$ for training and 2,875 for testing. During training, we randomly sample $2,048$ points on mesh surfaces. In the inference stage, we sample $2,048$ points multiple times to make sure all the points have at least ten predictions. We present the instance average IoU (mIoU, \%) and class average IoU (mcIoU, \%) in Table ~\ref{tb:part}. We observe that our \netseg\ achieves competitive or even better scores compared with existing methods, including PointNet, PointNet++, PCNN~\cite{atzmon2018point}, PointCNN, DGCNN, RSCNN~\cite{liu2019relation}, DensePoint~\cite{liu2019densepoint}, KPConv, 3D-GCN~\cite{3dgcn}, PAConv~, AdaptConv, and PointTransformer.
We can also find that the performance is improved slightly over the recent years. 
Specifically, while our model ranks third place w.r.t mcIoU, it yields higher mIoU (86.8\%) than previous approaches. By exploiting our proposed module to improve previous point cloud operations, notable improvements can be achieved. Detailed scores for each class are provided in the Supplementary Material.

\begin{table}[!t]
\centering
\caption{Part Segmentation Performance on ShapeNetPart}
\begin{threeparttable}
\setlength{\tabcolsep}{1.98mm}{
\begin{tabular}{ccc}
\toprule
{Method} & 
{mcIoU} & 
{mIoU} \\
\midrule
PointNet &
80.4 & 83.7\\
PointNet++  &
81.9 &85.1\\
PCNN  &
81.8 &85.1\\
PointCNN & 
84.6 &86.1\\
DGCNN & 
82.3 &85.1 \\
RSCNN & 
84.0 &86.2\\
DensePoint & 
84.2 & 86.4\\
KPConv & 
85.1&86.4\\
3D-GCN  & 82.7 & 85.3 \\
PAConv & 84.6 & 86.1  \\ 
AdaptConv & 83.4 &86.4\\
PointTransformer & 83.7 & 86.6 \\
\midrule
\multicolumn{3}{c}{Ours} \\
\midrule
Baseline & 84.0 & 86.5 \\
+\module & 84.3 & 86.6 \\
+\smodule* & 84.4 & 86.8 \\
RSConv & 83.1 & 86.0 \\
+\smodule & 84.1 & 86.4 \\
GAConv & 81.9 & 85.3 \\
+\smodule & 83.3 & 86.1 \\
RandLA & 83.5 & 86.2 \\
+\smodule & 83.8 & 86.4\\
\bottomrule
\end{tabular}
}
\begin{tablenotes}
    \item \textit{
    +\smodule* indicates our final model \netseg; Baseline indicates the model without \module\ or \smodule.}
\end{tablenotes}

\end{threeparttable}

\label{tb:part}
\end{table}

\begin{table} 
\centering
\caption{Shape Classification Performance on ModelNet40}
\begin{threeparttable}
\begin{tabular}{ccccc}
\toprule
Method &  Input & \#Points & mA & oA  \\ 
\midrule
PointNet++\ & PN & 5k & - & 91.9  \\
PointASNL & PN & 1k & - & 93.2 \\
PointTransformer & PN & 1k & 90.6 & 93.7 \\
PointNet &  P & 1k & 86.2 & 89.2  \\
PointNet++ & P & 1k & - & 90.7  \\
PointCNN & P & 1k & 88.1 & 92.2  \\

DGCNN~ & P & 2k & 90.7 & 93.5  \\
PointCNN &  P & 1k & 88.8 & 92.5  \\
DGCNN &  P & 1k & 90.2 & 92.9  \\
RSCNN & P & 1k & - & 93.6  \\
ShellNet &  P & 1k & - & 93.1 \\
PointASNL &  P & 1k & - & 92.9  \\
GridGCN&  P & 1k & 91.3 & 93.1  \\
PAConv & P & 1k & - & 93.9  \\
AdaptConv & P & 1k & 90.7 & 93.4 \\
\midrule
\multicolumn{5}{c}{Ours}\\
\midrule
Baseline & P & 1k & 90.7&93.5 \\
+\module & P & 1k & 91.0&93.9 \\
+\smodule* &  P & 1k & 91.6 & 94.2  \\
RSConv & P & 1k & 90.0 & 93.2 \\
+\smodule & P & 1k & 90.4 & 93.3 \\
GAConv  & P & 1k &89.2 &92.7 \\
+\smodule  & P & 1k & 89.7&93.5\\
RandLA  & P & 1k & 89.3 & 92.9\\
+\smodule & P & 1k & 89.9 & 93.1\\
\bottomrule
\end{tabular}
\begin{tablenotes}
    \item \textit{P denotes Point, while PN denotes Point and Normal.
    +\smodule* indicates our final model \netcls; Baseline indicates the model without \module\ or \smodule.}
\end{tablenotes}

\end{threeparttable}
\label{tab:modelnet40}
\end{table}

\begin{table} \footnotesize
\centering
\caption{Shape Classification Performance on Three Subsets of ScanObjectNN}
\begin{threeparttable}
\begin{tabular}{cccc}
\toprule
Method &  OBJ-BG & OBJ-ONLY & PB-T50-RS \\ 
\midrule
PointNet & 73.3 & 79.2 & 68.2  \\
PointNet++ & 82.3 &84.3 &77.9 \\
DGCNN & 82.8 & 86.2 & 78.1\\
PointCNN & 86.1 & 85.5 & 78.5 \\
\midrule
\multicolumn{4}{c}{Ours}\\
\midrule
Baseline &  90.5 &  90.7 & 84.4 \\
+\module & 91.0 & 91.2 & 86.4 \\
+\smodule* &  91.2 & 91.4 & 86.9  \\
RSConv & 87.6 & 88.0 & 82.3\\
+\smodule & 90.7 & 88.6 & 84.5 \\
GAConv  & 85.7 & 85.5 & 80.7 \\
+\smodule  & 87.3 & 88.1 &83.1\\
RandLA  & 86.7 & 86.6 & 83.4\\
+\smodule & 88.3 & 88.5 &  83.7\\
\bottomrule
\end{tabular}
\begin{tablenotes}
    \item \textit{
    +\smodule* indicates our final model \netcls; Baseline indicates the model without \module\ or \smodule.}
\end{tablenotes}
\end{threeparttable}
\label{tab:sonn}
\end{table}

\subsection{3D Shape Classification}
\label{subsec:modelnet40}

We further evaluate our model on two 3D shape classification datasets, including ModelNet40~\cite{wu20153d} containing $9,843$ 3D CAD models for training and $2,468$ for testing and ScanObjectNN~\cite{scanobjectnn} containing around $2,900$ real-world point clouds. 
For ModelNet40, during training, we uniformly sample $1,024$ points on the mesh surfaces. We make comparisons against previous methods: PointNet, PointNet++, DGCNN, PointCNN, PointASNL~\cite{yan2020pointasnl}, RSCNN, GridGCN~\cite{xu2020grid}, ShellNet~\cite{zhang2019shellnet}, PAConv, PointTransformer, and AdaptConv. For ScanObjectNN, we follow previous works and conduct experiments on three subsets: OBJ-BG, OBJ-ONLY, and PB-T50-RS, and compare our method with PointNet, PointNet++, DGCNN, and PointCNN. The comparisons in the two tables can show the capability of our method on the shape classification task.

\begin{table} \footnotesize
\centering
\caption{Ablation Study on The Proposed Modules
}
\begin{threeparttable}
\setlength{\tabcolsep}{2.4mm}{
\begin{tabular}{ccccc}
\toprule
Analysis & Config. & oA & mAcc & mIoU \\
\midrule
\multirow{4}{*}{} & $K_e=2$ & 89.3 & 73.0  & 66.2  \\
{\#Nearest}&$K_e=3$ & 89.8 & 73.1 & 66.8  \\
{Neighbours}& $K_e=4$ &89.7 &73.5 &67.3 \\
&$K_e=5$ & 89.4  & 73.1 & 66.8  \\
\midrule

\multirow{3}{*}{Effectiveness} &AFA* & 87.9 & 69.8 & 63.9 \\
&\module* & 88.1 & 70.0 & 64.1 \\
{of modules}&\smodule* & 88.1 & 70.6 & 64.2 \\
\midrule
\multirow{4}{*}{} & $D^f$ & 87.9 & 70.7 & 64.2 \\ 
Relation&$D^f+D^s$ & 89.1 & 72.8 & 65.9 \\ 
learning&$D^f+D^c$ & 88.6  & 71.6 & 65.5 \\
&$D^f+D^c+D^s$ &89.7 & 73.5 &67.3 \\
\bottomrule
\end{tabular}
}
\begin{tablenotes}
    \item \textit{
    The details are provided in the text (Sec.~\ref{sec:ablation}).}
\end{tablenotes}
\end{threeparttable}

\label{tab:ab}
\end{table}
\subsection{Ablation Study}
\label{sec:ablation}
The comparisons against the state-of-the-art methods demonstrate the effectiveness of our model in exploiting the local structures.
Here, we conduct extensive ablations to inspect the proposed modules on Area 5 of S3DIS~\cite{armeni20163d} and report the results (no voting scheme used) in Table~\ref{tab:ab}.
\newline

\noindent \textbf{Number of nearest neighbours.} We first study the impacts of the number of nearest neighbors for an edge, \textit{i.e.,} $K_e$. As reported in Table~\ref{tab:ab}, the model performs better when we set $K_e$ to 4, which is used in all experiments. It is also worth noting that when we set $K_e$ to 2 or 3, the model still outperforms most of existing works list in Table~\ref{tab:s3dis}. When setting $K_e$ to 5, we find that the performance drops, due to the aggregation of some unrelated information.
\newline

\noindent \textbf{Effectiveness of the proposed modules.} 
In previous sections, we provide the performance of the baseline model (no edge-to-edge interaction), the model with the basic edge-to-edge interaction and the full model on all tasks. We also study the proposed module by combining it with previous point cloud analysis operations. These comparisons are able to show the effectiveness of our method. Here, to further provide supports for the analysis on the differences between ours and PointWeb experimentally, we re-implement the AFA module proposed by PointWeb in our framework (AFA*). The comparisons between AFA* and the simplified version of our modules (\module* and \smodule*) in Table~\ref{tab:ab} also show the superiority of our methods over AFA strategy in PointWeb.
\newline

\noindent \textbf{Relations learning between edges.} In Eq.~\ref{eq:gamma}, we introduce three kinds of information, \myie{relative position $D^s$, difference between features $D^f$, and normal vector $D^c$}, to learn the low and high-level relations between the edges. To evaluate their impacts, we revise \smodule\ by considering four kinds of feature combinations, including $D^f$, $D^f+D^s$, $D^f+D^c$, and $D^f+D^s+D^c$.
The results in Table~\ref{tab:ab} show that the geometric relations $D^s$ and $D^c$ can both bring improvements, and when all relations are exploited, the performance can be improved greatly.
\newline


\begin{table}[htp]\footnotesize
\centering
\caption{
Comparisons of Performance and Model Complexity}
\begin{threeparttable}
\setlength{\tabcolsep}{0.4mm}{
\begin{tabular}{cccccc}
\toprule
Method & mIoU (A5 / 6F) & \#Par./M & FLOPs/G & Mem./G & T./s \\
\midrule
PointWeb & 60.3 / 66.7 & 1.0 & 142 & 8.4 & 0.18 \\
FPConv & 62.8 / 68.7 & 17.4 & 1032 & 9.8 & 0.69 \\
PAConv & 66.6 / 69.3 & 0.6 & 52 & 13.2 & 0.28 \\
AdaptConv & 67.9 / - & 15.8 & 492 & 15.6 & 0.81\\
Ours & 68.0 / 73.7 & 3.6 & 312 & 9.4 & 0.37 \\
\bottomrule
\end{tabular}
}
\begin{tablenotes}
    \item \textit{
   We provide the
mIoUs (Area 5 and 6-fold), the parameters (\#Par.), FLOPs, inference memory (Mem.), and inference time (T.) of the segmentation models on S3DIS dataset.
    We calculate the FLOPs, Mem., and T. by processing 12 samples, each one containing $14,000$ points, on one Tesla v100 GPU.}
\end{tablenotes}
\end{threeparttable}
\label{tb:param}
\end{table}
\begin{table}[htp]
\centering
\caption{Robustness Analysis}
\begin{threeparttable}
\setlength{\tabcolsep}{0.6mm}{
\begin{tabular}{ccccccccc}
\toprule
Method & None & $90^\circ$ & $180^\circ$ & $270^\circ$ &$\times0.8$ & $\times1.2$ & $0.5\%$  & $1\%$ \\
\midrule
PointWeb & 54.7 & 52.5 & 54.2 & 51.5 & 54.0 & 51.7 & 54.5 & 54.4 \\
FPConv &  56.0 & 54.0 & 54.3 & 52.4 & 54.3 & 52.5 & 55.6 & 55.1 \\
Ours* &  58.1 & 56.5 & 58.0 & 55.6 & 58.0 & 56.3 & 57.9 & 57.7 \\
\midrule
PAConv & 59.5 & 55.4 & 58.1 & 53.9 & 58.7 & 59.1 & 58.8 & 58.3  \\
Ours & 62.3 & 62.1 & 59.9 & 61.1 & 62.3 & 59.4 & 62.0 & 61.2 \\
\bottomrule
\end{tabular}
}
\begin{tablenotes}
    \item \textit{
    We evaluate the robustness through performing rotation ($90^\circ$,$180^\circ$,$270^\circ$), scaling ($\times0.8$,$\times1.2$), and adding noises ($0.5\%,1\%$) on 20 rooms of S3DIS dataset. Since FPConv and PointWeb are trained without data augmentation (DA), for fair comparisons, we also re-train our model without DA (Ours*).}
\end{tablenotes}
\end{threeparttable}
\label{tb:rob}
\end{table}

\subsection{Model Complexity} 
\label{sec:comp}
\modify{In Table~\ref{tb:param}, we report the mIoUs, parameters, FLOPs, inference memory, and inference time of our segmentation model and several state-of-the-art works, including 
PointWeb, FPConv, PAConv, and AdaptConv. We can observe that our model's complexities and running time are competitive to recent approaches, while our model outperforms or generates comparable results to previous methods for various point cloud analysis tasks.}

\subsection{Robustness Analysis}
Our method aims at enhancing the point-to-point relation through modeling the edge-to-edge interaction adaptively, which could make the relation aware of the local structure and thus more robust to the geometry transformation. We make robustness evaluation through performing rotation, scaling, and adding noises on the input data during inference. As shown in  Table~\ref{tb:rob}, our method performs better than previous methods under all settings.


\subsection{Visualization Results}

\begin{figure}[!t]
\begin{center}
\includegraphics[width=0.98\linewidth,height=2.0in]{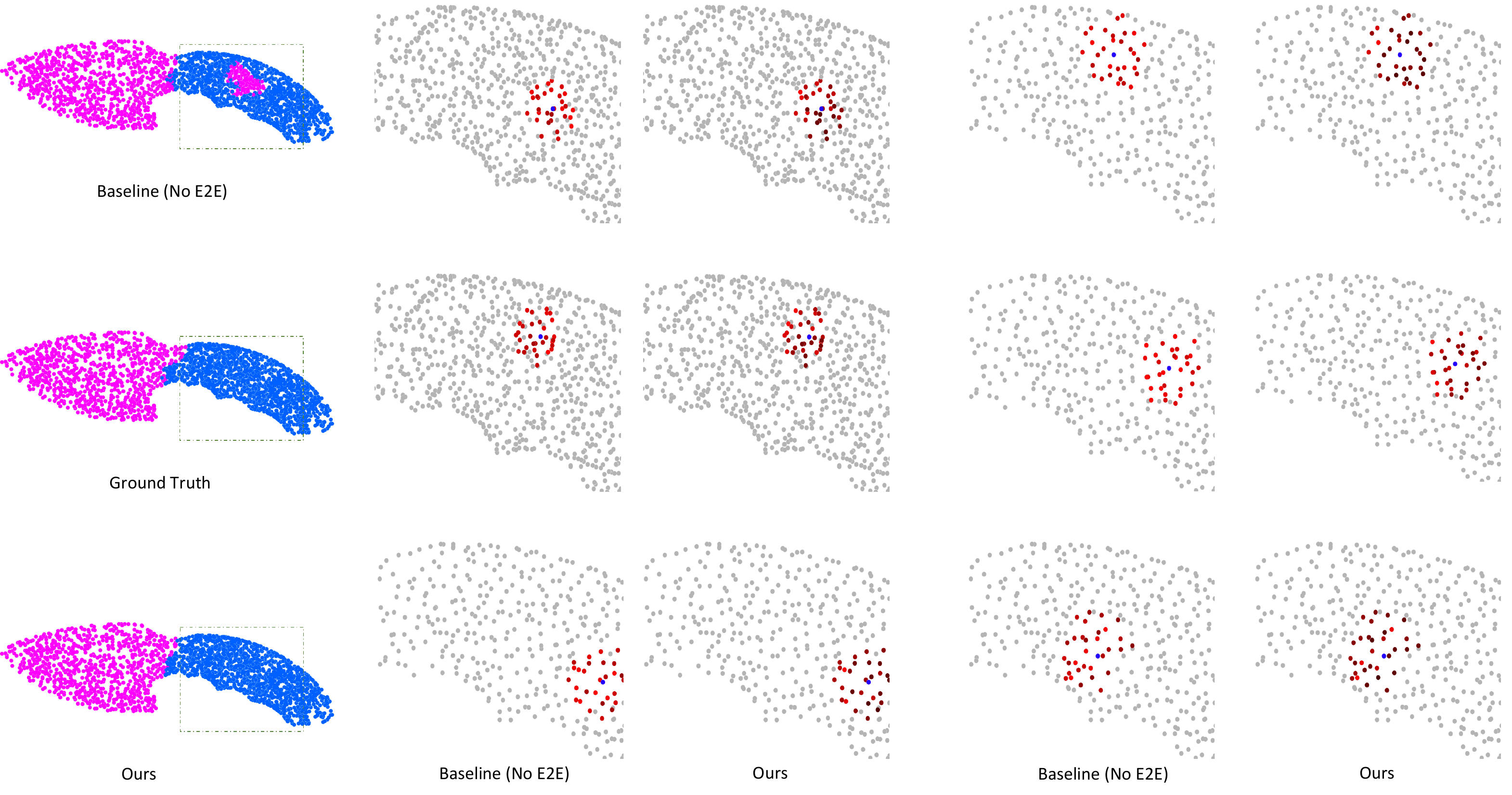}
\end{center}
\caption{Qualitative analysis on ShapeNetPart~\cite{Yi16}. Zoom in for more details (in color).}
\label{a:fig:ana_e2e}
\end{figure}

\begin{figure}[!t]
\centering
    \subfloat[]{\includegraphics[width=0.2\linewidth,height=0.7in]{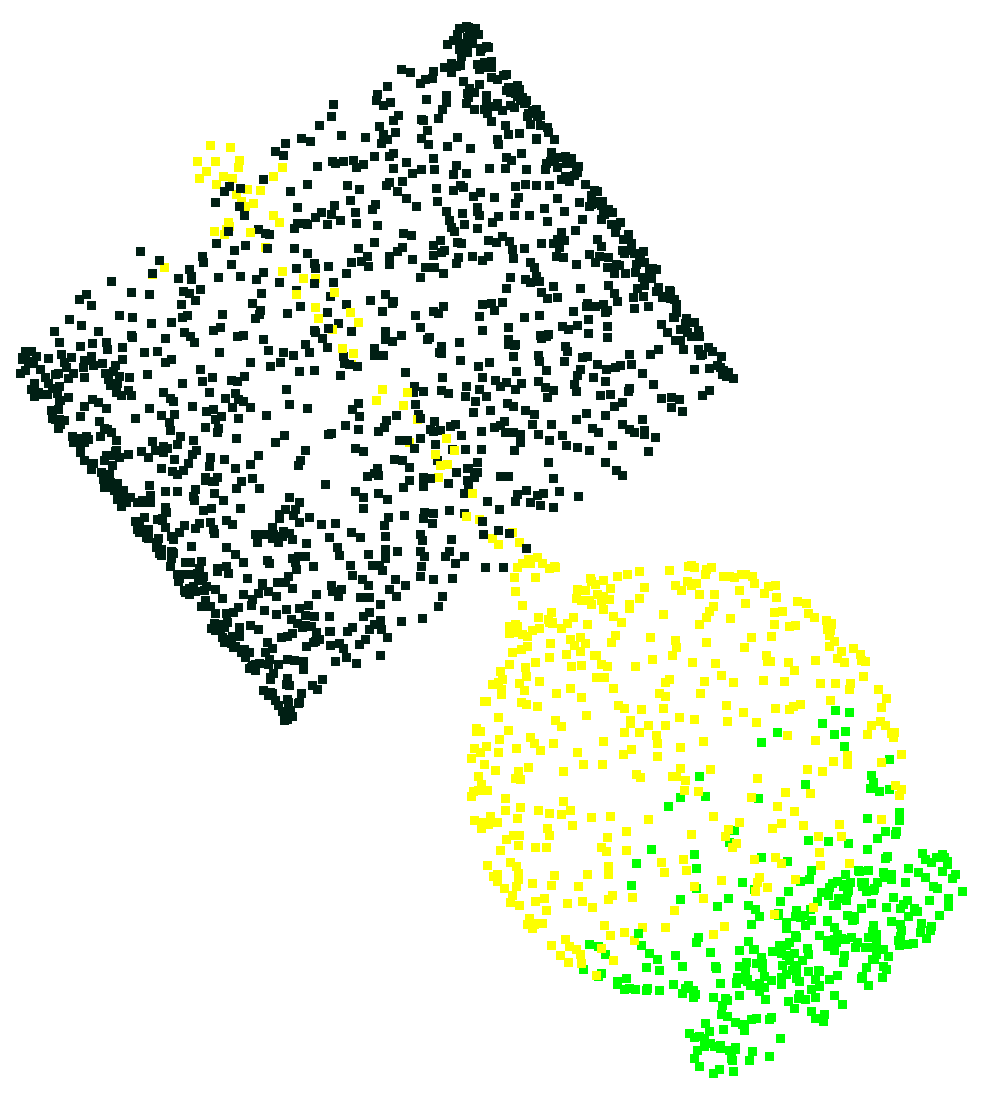}}
    \hfil
     \subfloat[]{ \includegraphics[width=0.2\linewidth,height=0.7in]{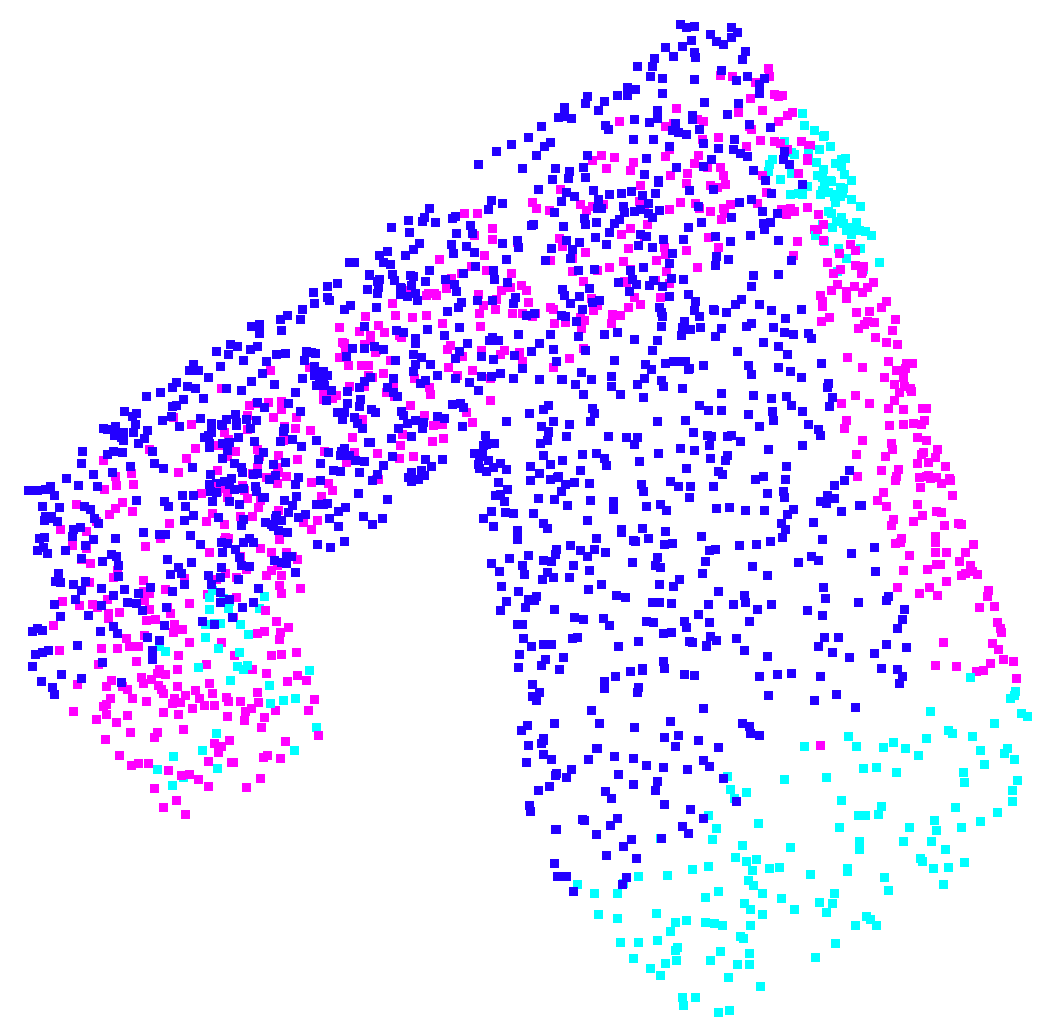}}\hfil
      \subfloat[]{\includegraphics[width=0.2\linewidth,height=0.7in]{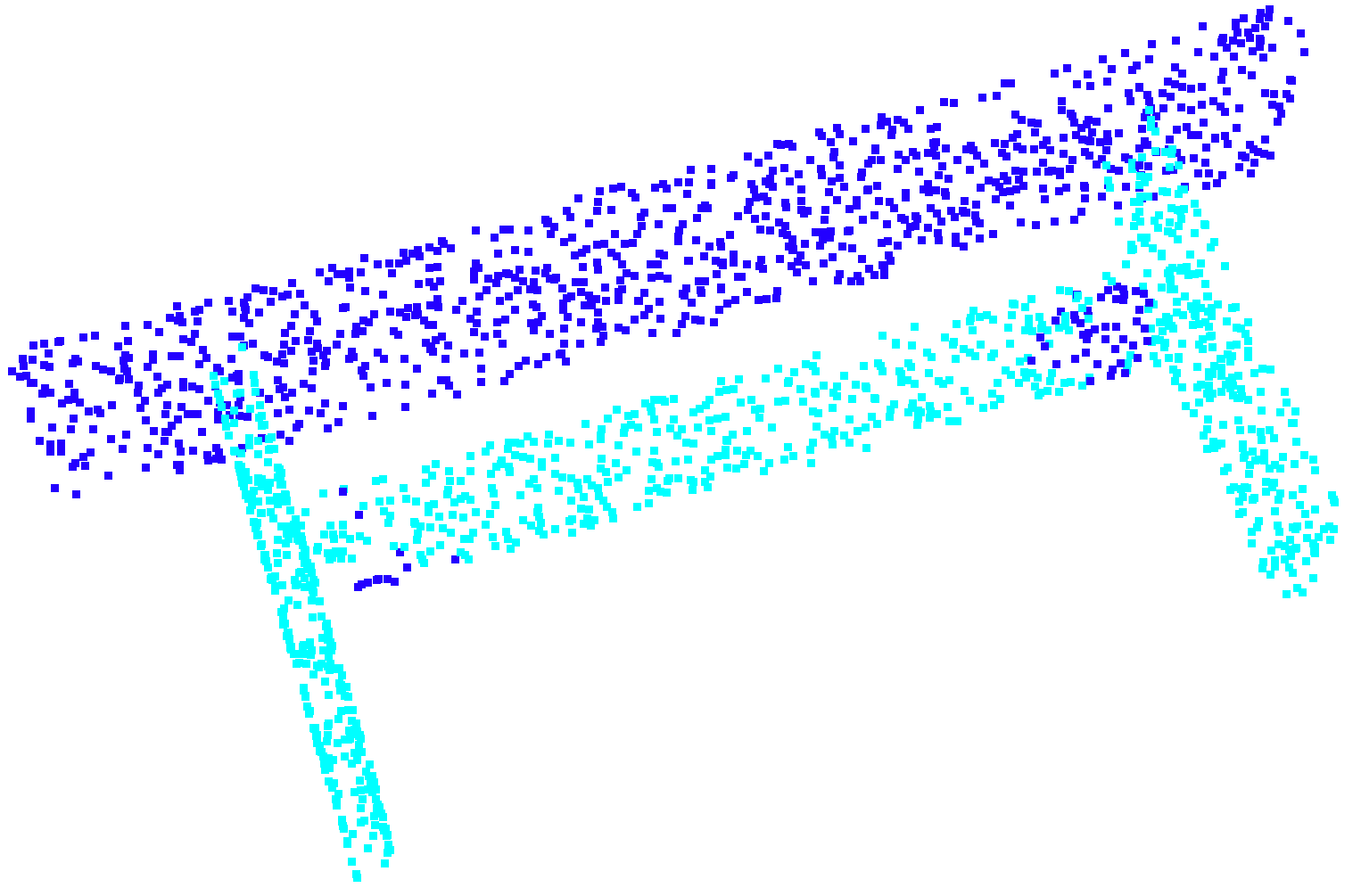}}
      \hfil
     \subfloat[]{ \includegraphics[width=0.2\linewidth,height=0.7in]{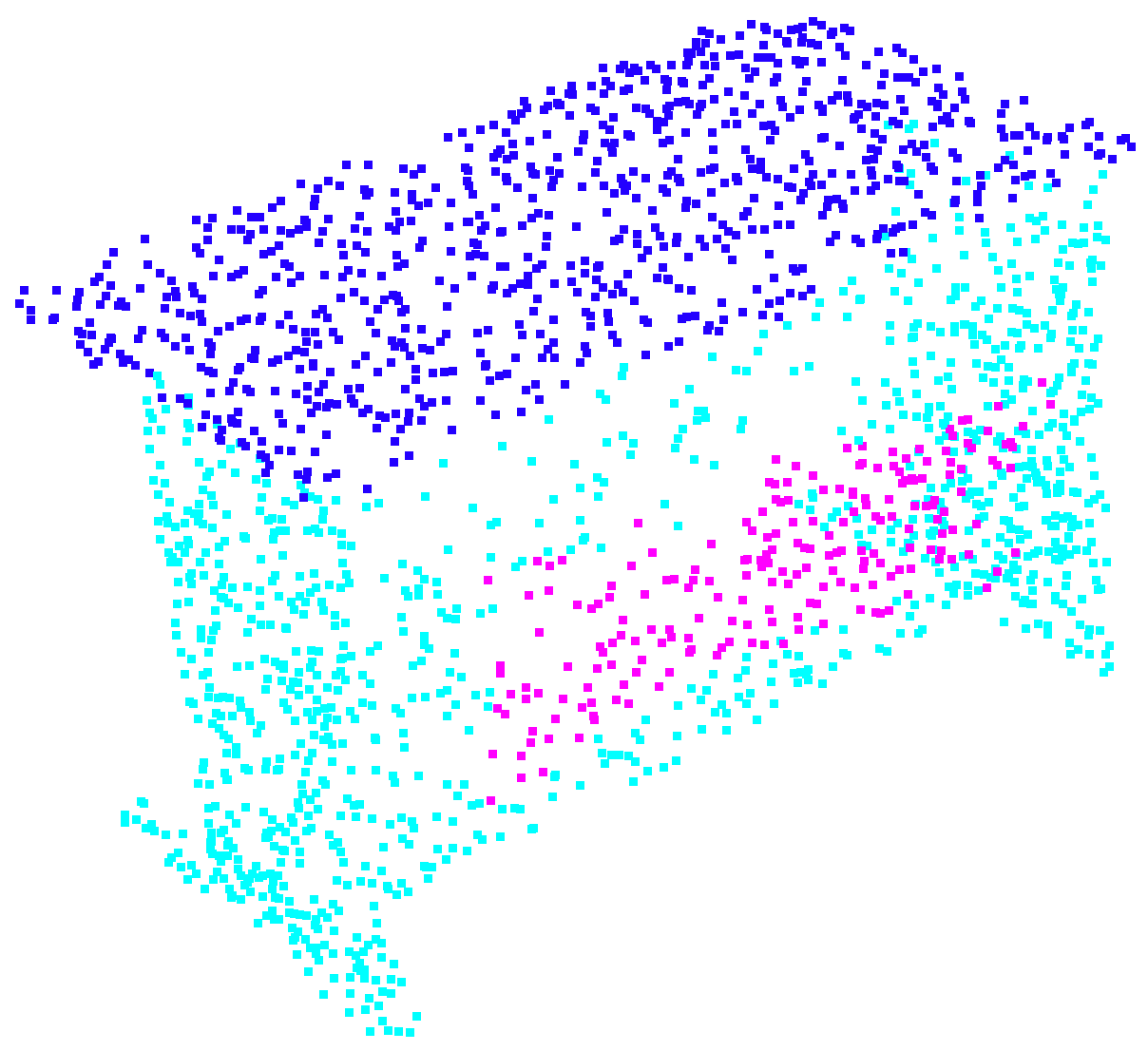}}
     \hfil
     \quad
    \subfloat[]{\includegraphics[width=0.2\linewidth,height=0.7in]{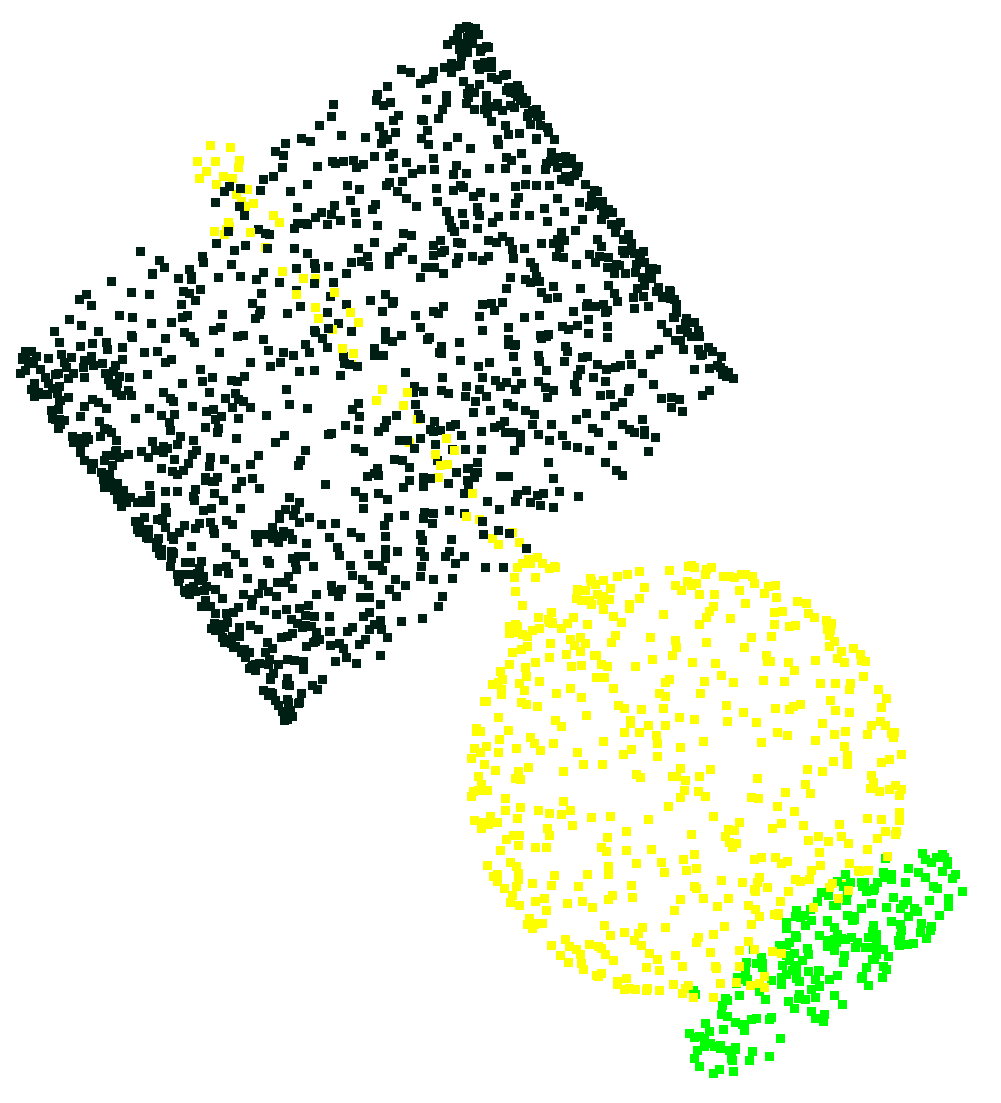}}\hfil
     \subfloat[]{ \includegraphics[width=0.2\linewidth,height=0.7in]{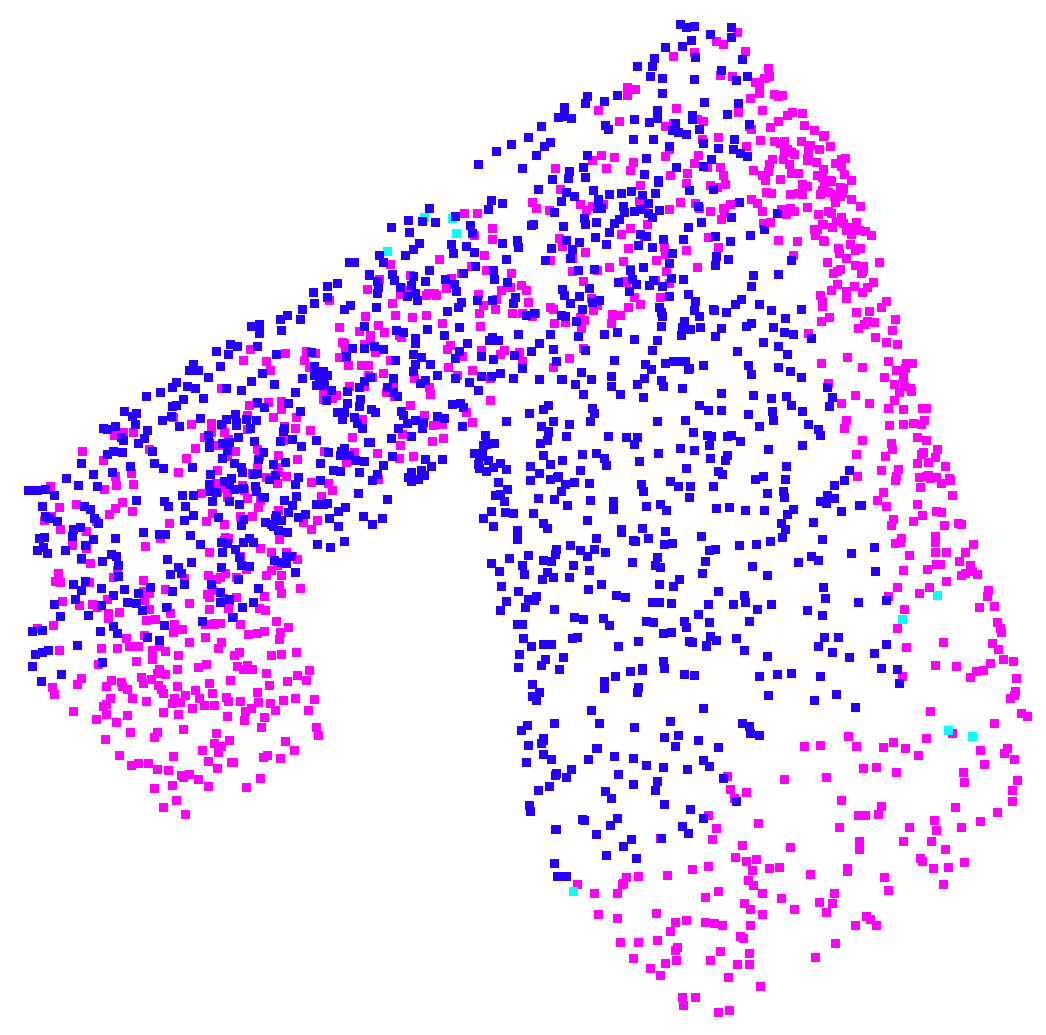}}\hfil
      \subfloat[]{\includegraphics[width=0.2\linewidth,height=0.7in]{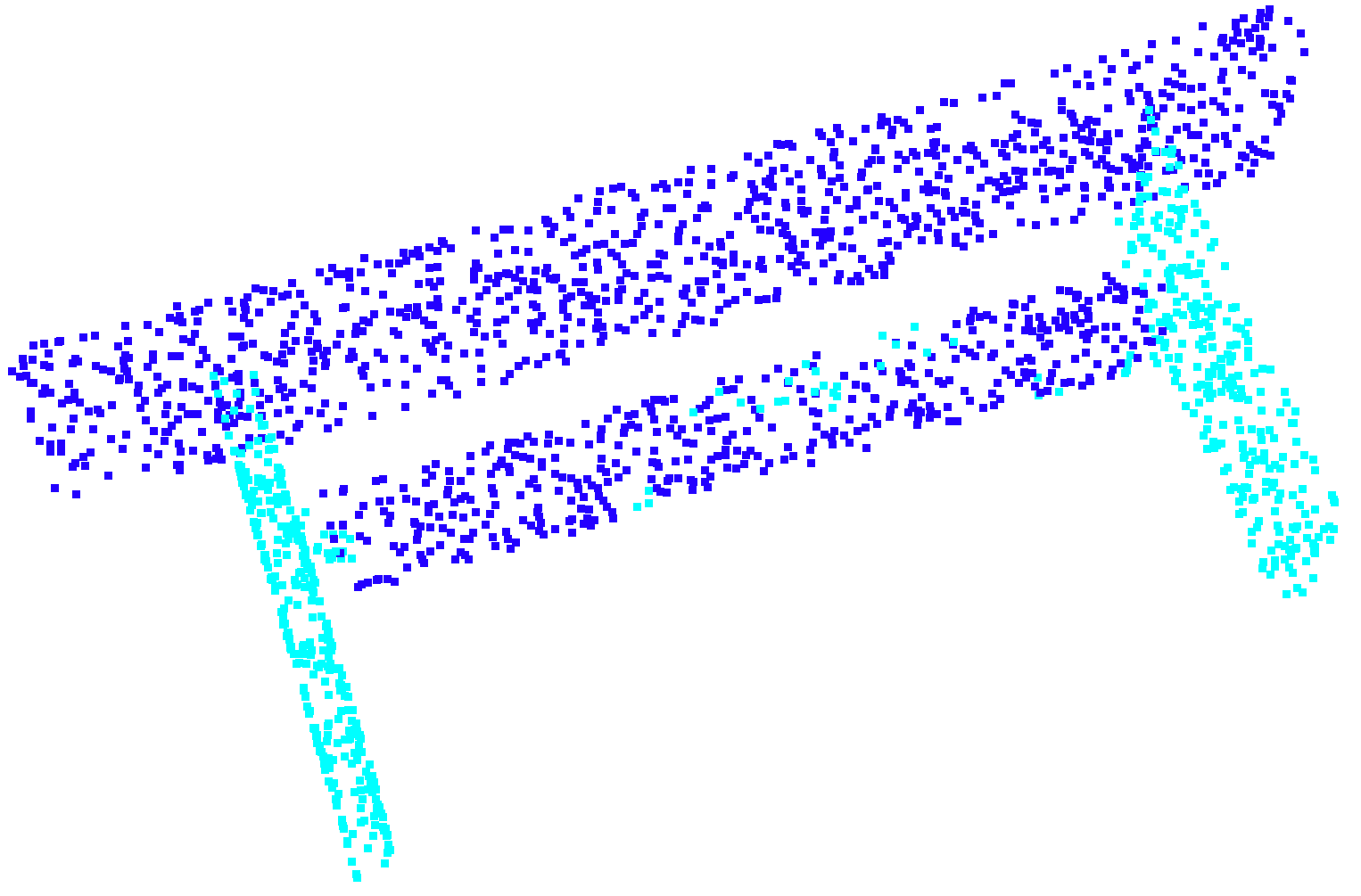}}\hfil
     \subfloat[]{ \includegraphics[width=0.2\linewidth,height=0.7in]{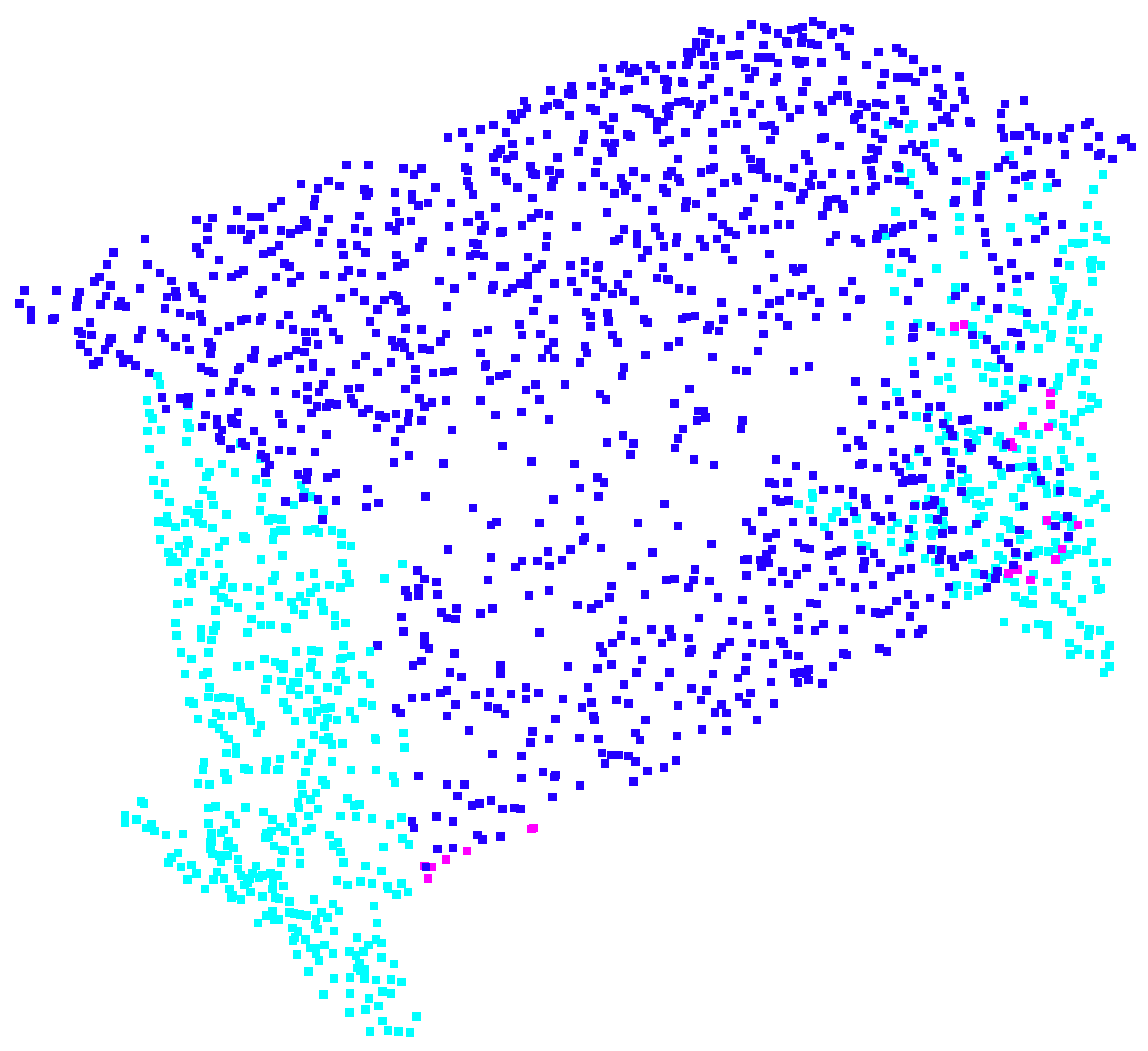}}\hfil
    \quad
    \subfloat[]{\includegraphics[width=0.2\linewidth,height=0.7in]{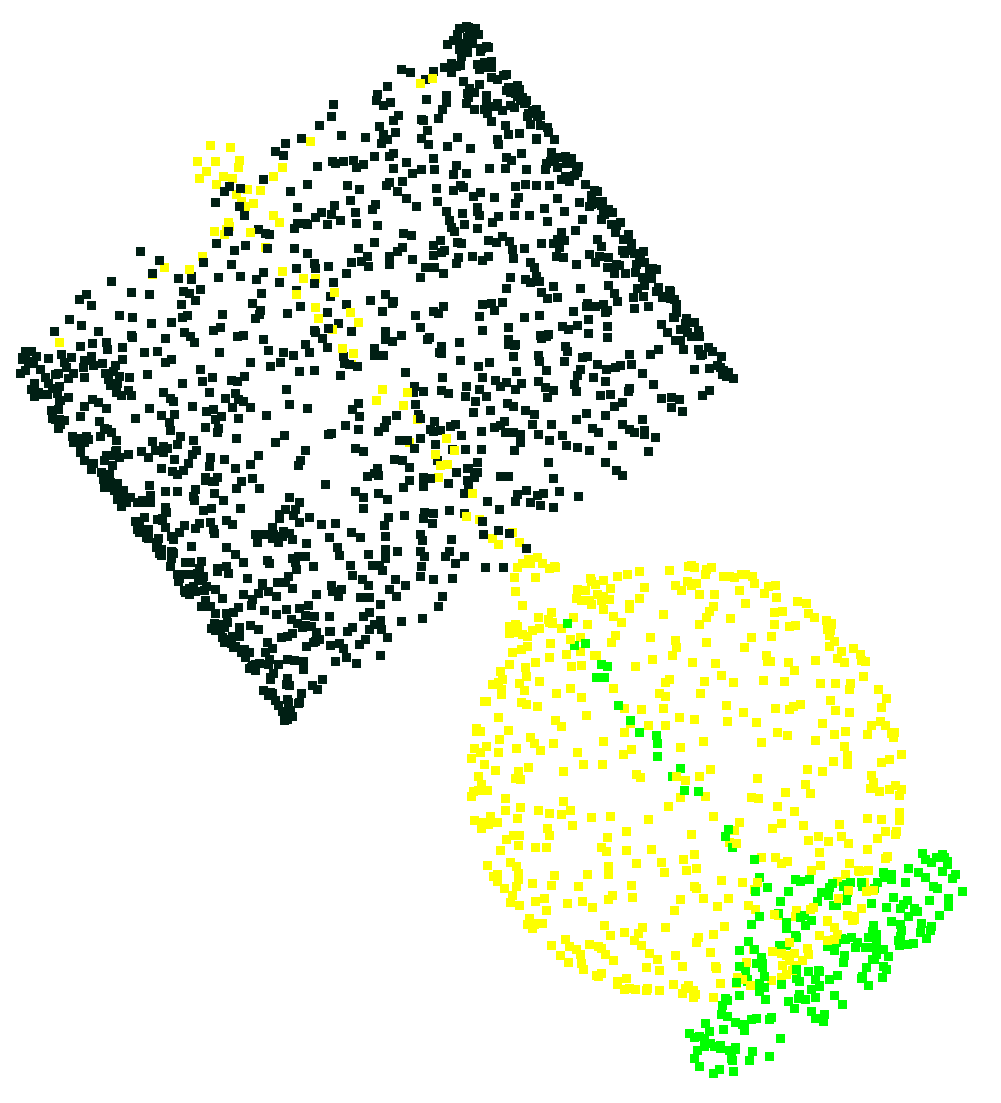}}\hfil
     \subfloat[]{ \includegraphics[width=0.2\linewidth,height=0.7in]{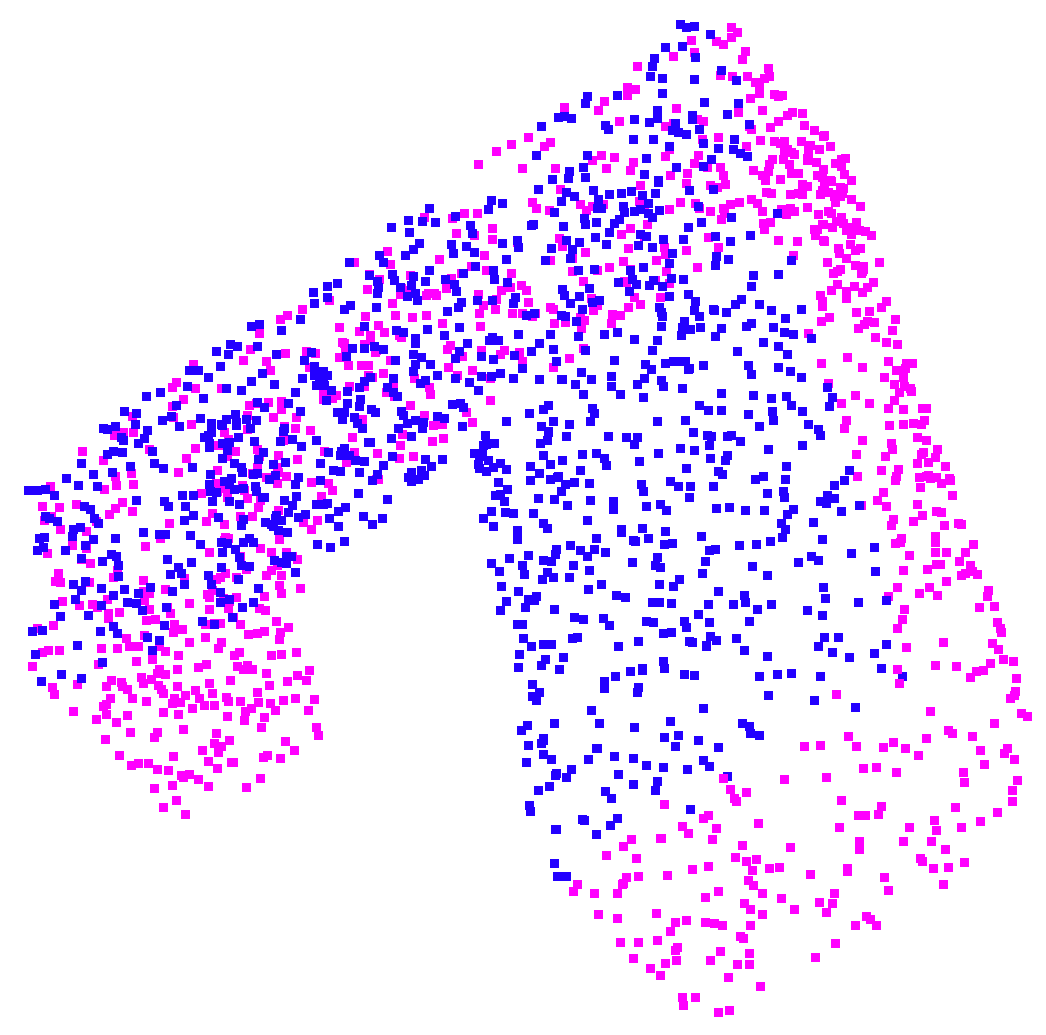}}\hfil
      \subfloat[]{\includegraphics[width=0.2\linewidth,height=0.7in]{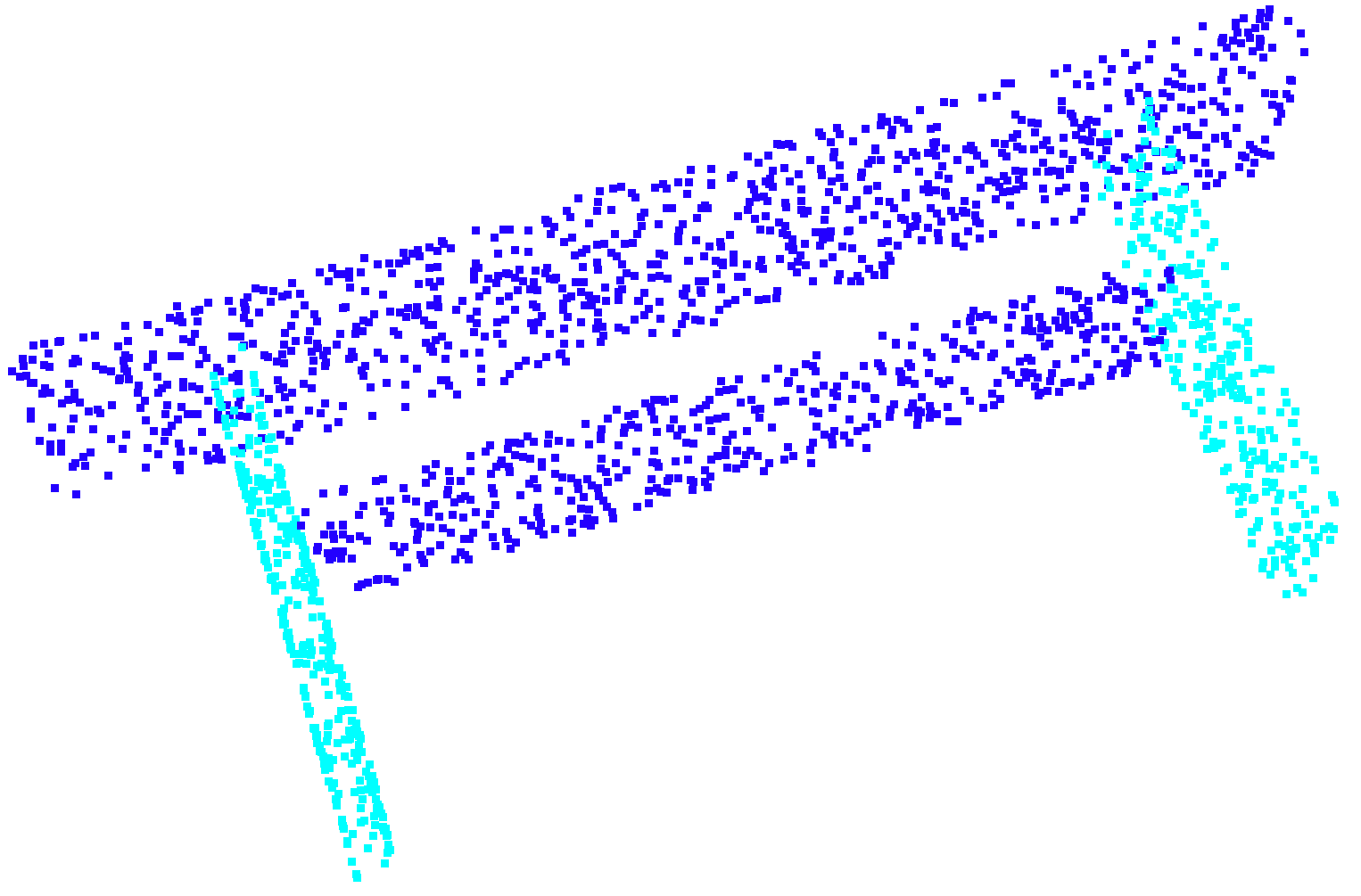}}\hfil
     \subfloat[]{ \includegraphics[width=0.2\linewidth,height=0.7in]{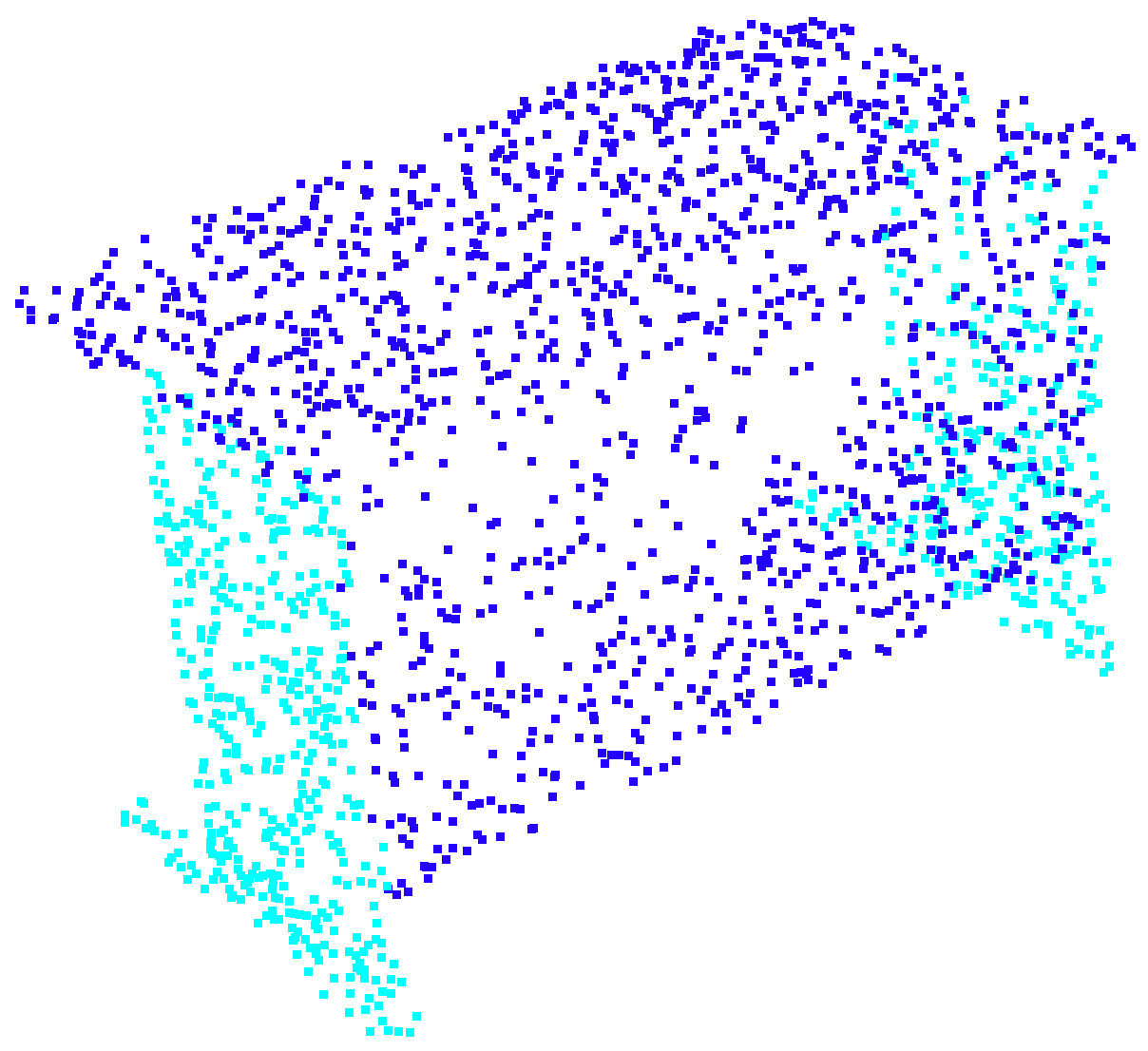}}\hfil
    
  \caption{Qualitative comparisons on ShapeNetPart~\cite{Yi16}. (a)-(d): Results of Baseline; (e)-(h): Results of Ours; (i)-(l): Ground Truth. Best viewed in color.}
  \label{a:fig:shaper_comp} 
\end{figure}

Our edge-to-edge interaction focuses on improving the point-to-point relation, which is learned from two adjacent points solely. In this way, the point-to-point relation is enabled to be shape-aware and thus to be more discriminative. For some unique structures, the relation between two adjacent points can provide enough information to model the local shape, while for the common structures, it is not enough to consider each point pair solely. We provide a brief illustration to assist in understanding this in Figure~\ref{a:fig:ana_e2e}. The left three images show that our model can predict the part category more accurately by introducing the edge-to-edge interaction. We also visualize the learned relation with / without edge-to-edge interaction at different scales by using different shades of red color to represent the relation between the corresponding neighbouring point and the central point (blue). As shown in Figure~\ref{a:fig:ana_e2e}, we can observe that 
the colors of the neighbours in Baseline are close, while the colors in ours are more distinct. It means that
our method is able to adjust the relation adaptively and exploit the neighbours more flexibly. More visualization comparisons can be found in Figure~\ref{a:fig:shaper_comp}.

\begin{figure*}

\centering
      \subfloat{\includegraphics[width=0.2\linewidth,height=1in]{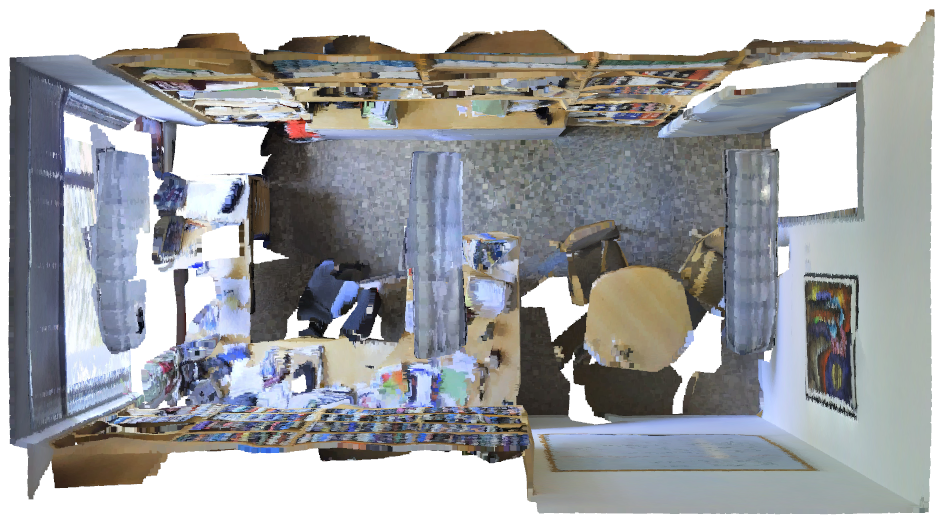}}\hfil
      \subfloat{\includegraphics[width=0.2\linewidth,height=1in]{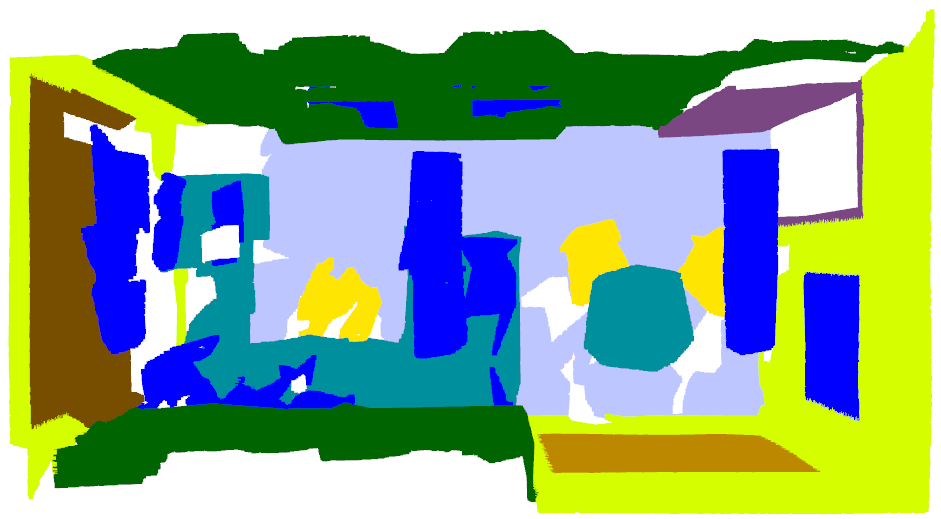}}\hfil
      \subfloat{\includegraphics[width=0.2\linewidth,height=1in]{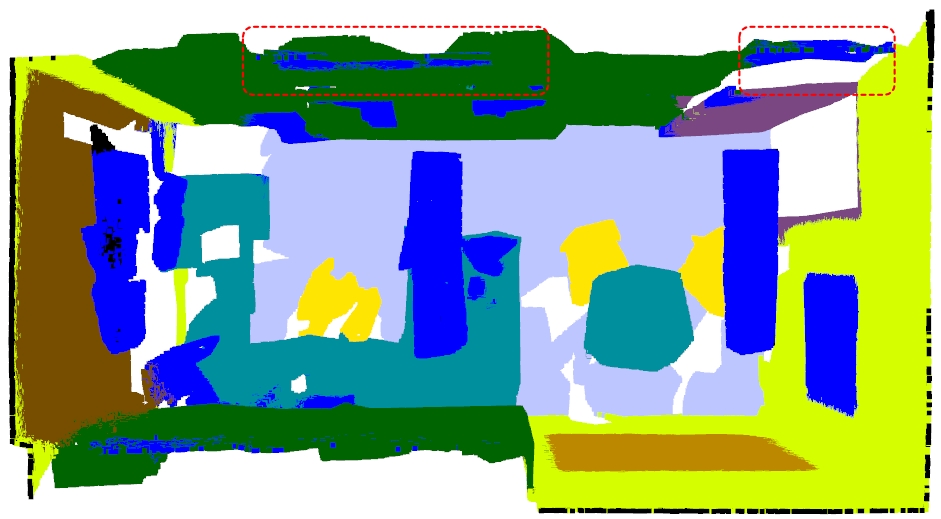}}\hfil
      \subfloat{\includegraphics[width=0.2\linewidth,height=1in]{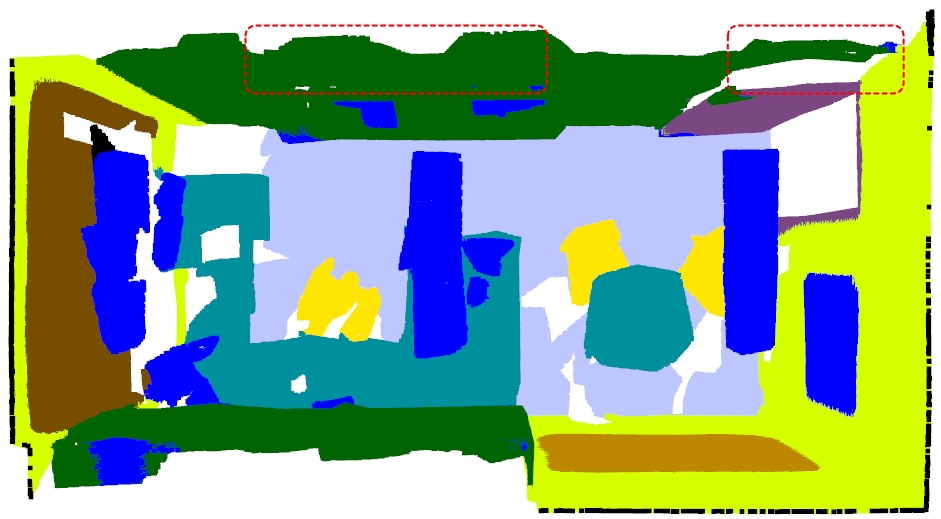}}\hfil
      
      
      \subfloat{\includegraphics[width=0.22\linewidth,height=1in]{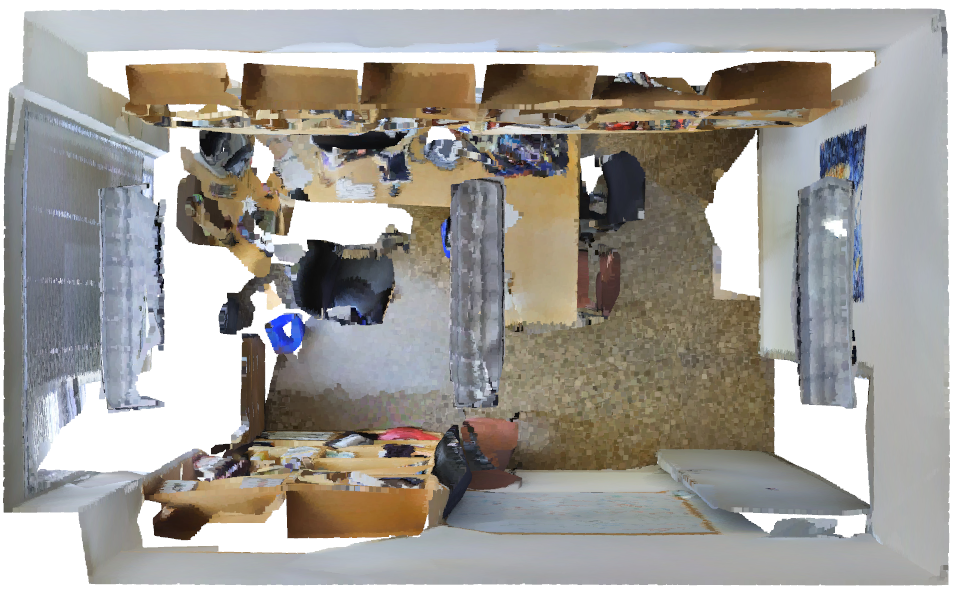}}\hfil
       \subfloat{ \includegraphics[width=0.22\linewidth,height=1in]{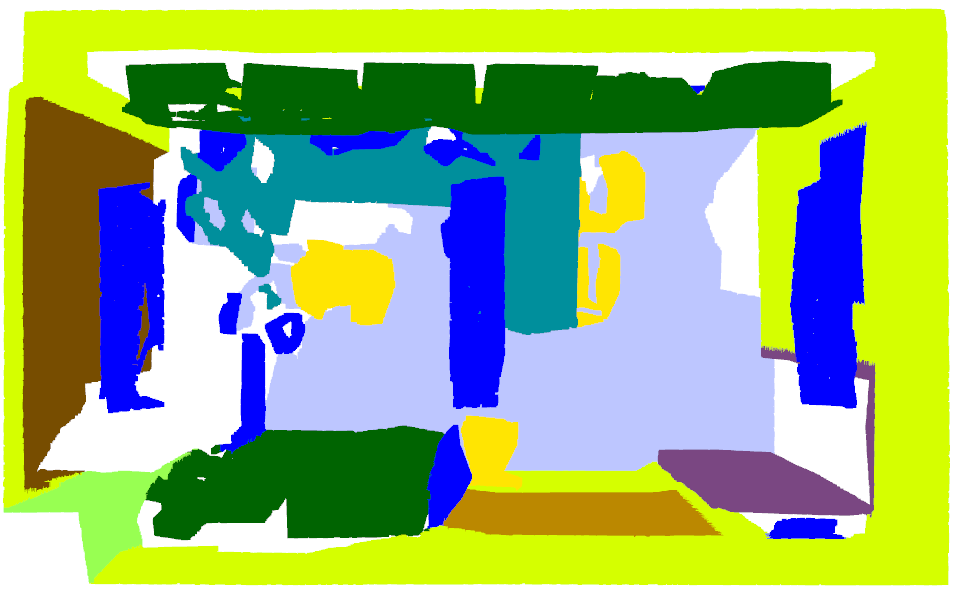}}\hfil
      \subfloat{\includegraphics[width=0.22\linewidth,height=1in]{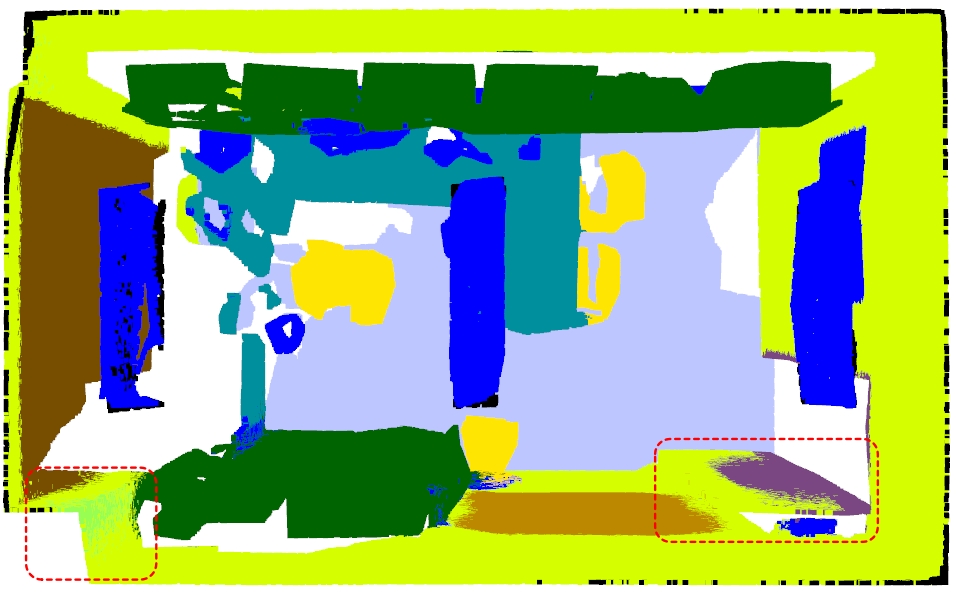}}\hfil
      \subfloat{  \includegraphics[width=0.22\linewidth,height=1in]{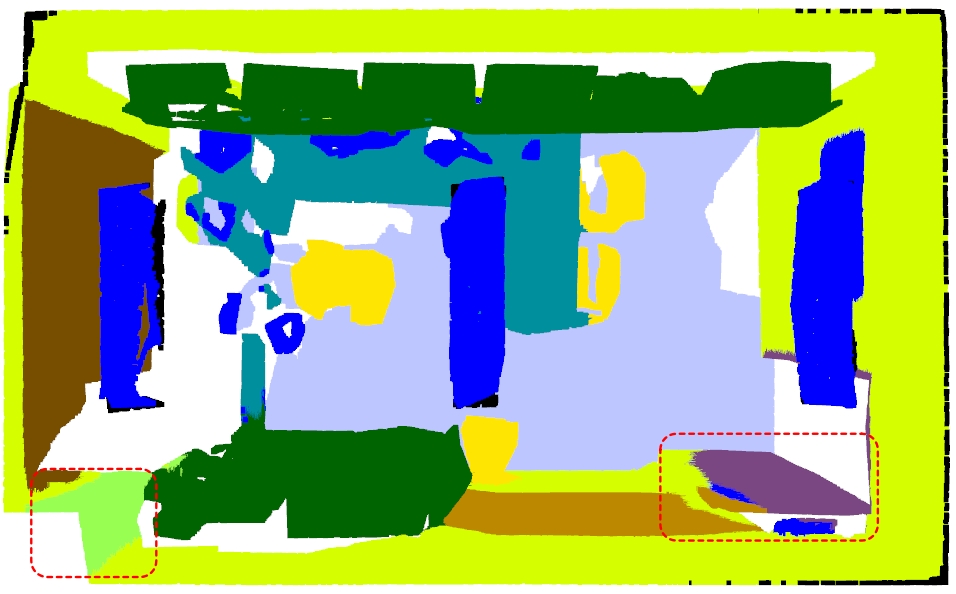}}\hfil
      
      \subfloat{\includegraphics[width=0.22\linewidth,height=1in]{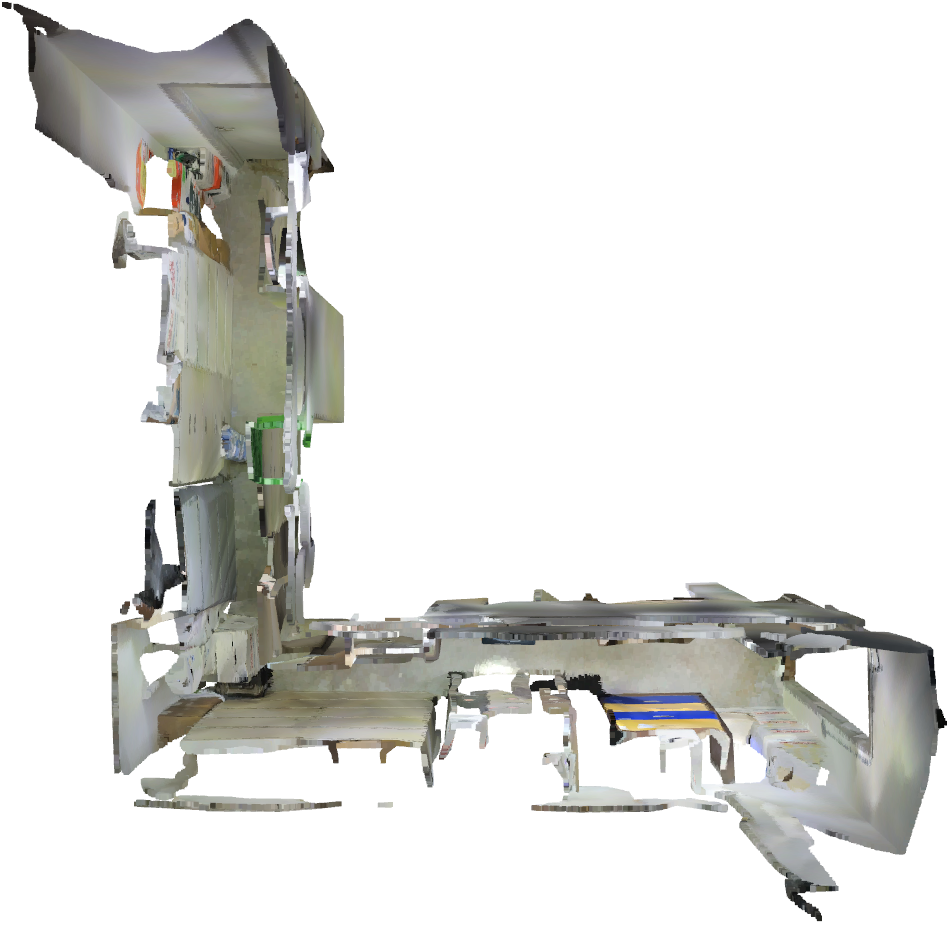}}\hfil
      \subfloat{\includegraphics[width=0.22\linewidth,height=1in]{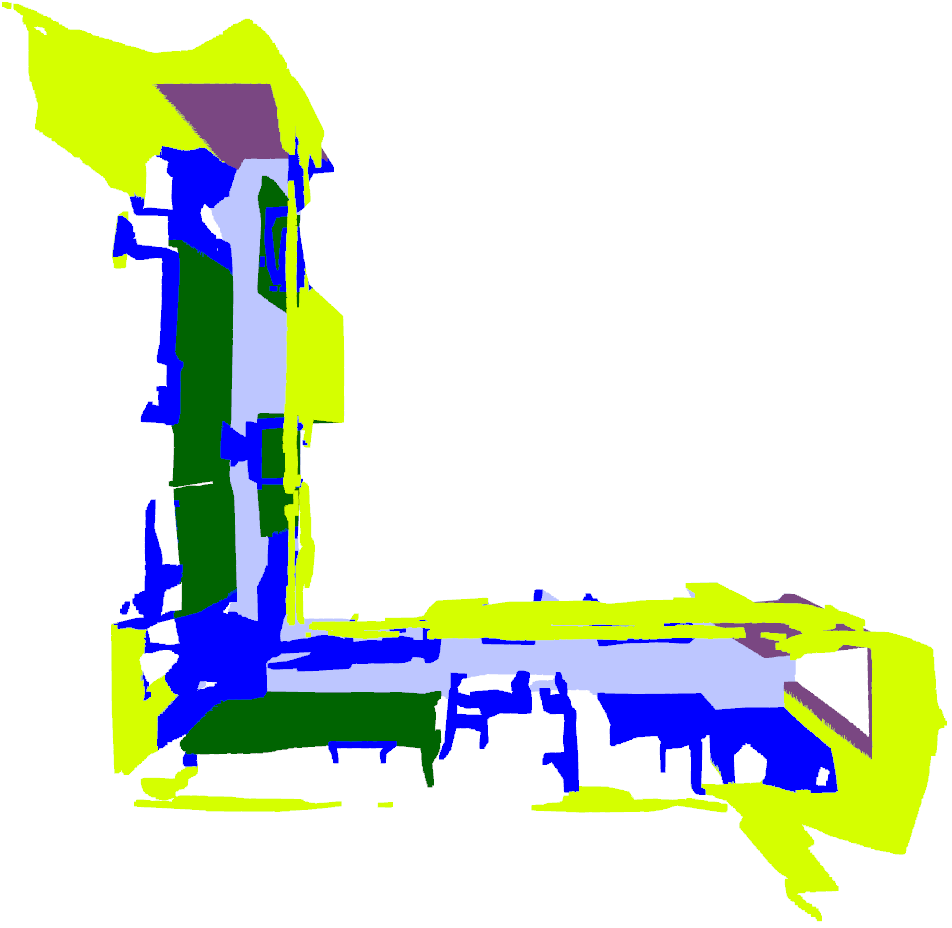}}\hfil
      \subfloat{\includegraphics[width=0.22\linewidth,height=1in]{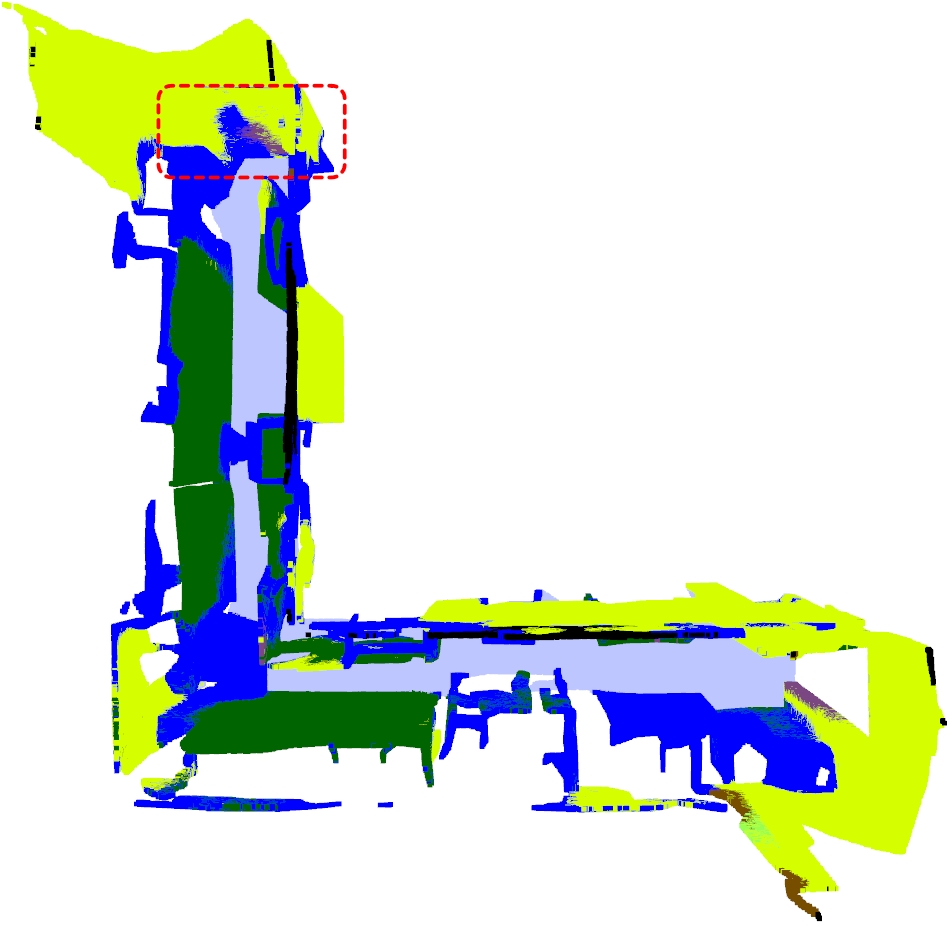}}\hfil
      \subfloat{  \includegraphics[width=0.22\linewidth,height=1in]{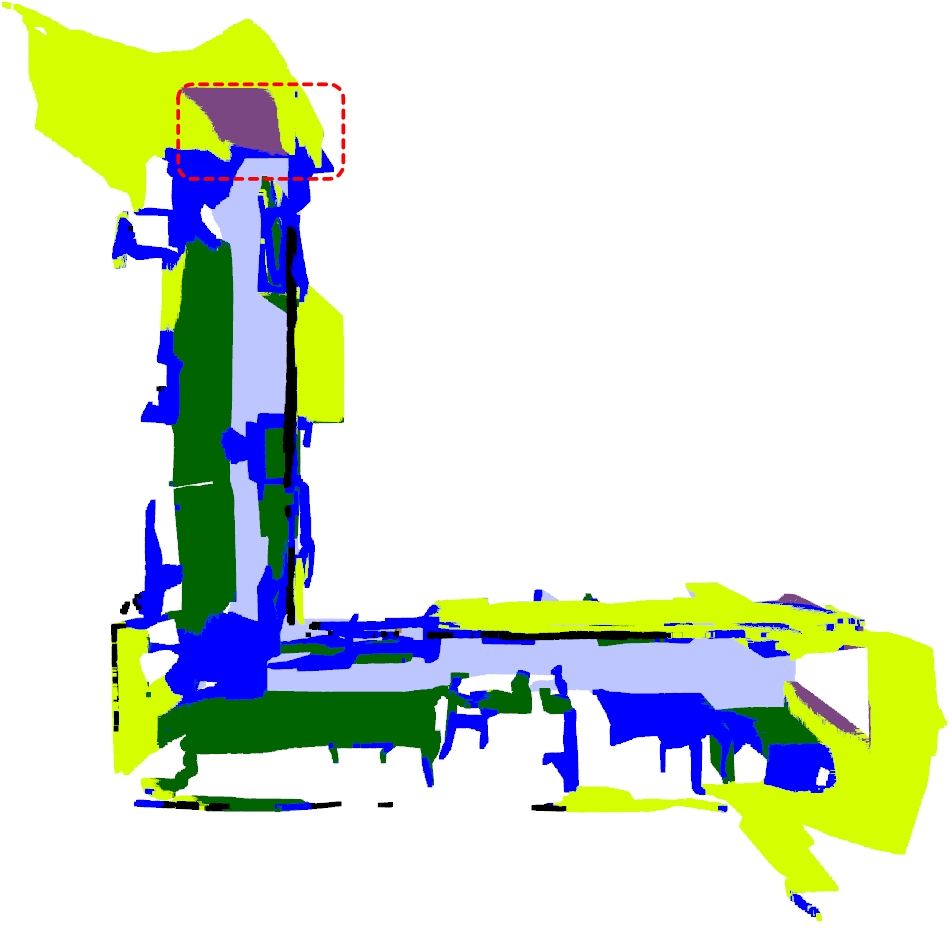}}\hfil
      
      \subfloat{\includegraphics[width=0.22\linewidth,height=1in]{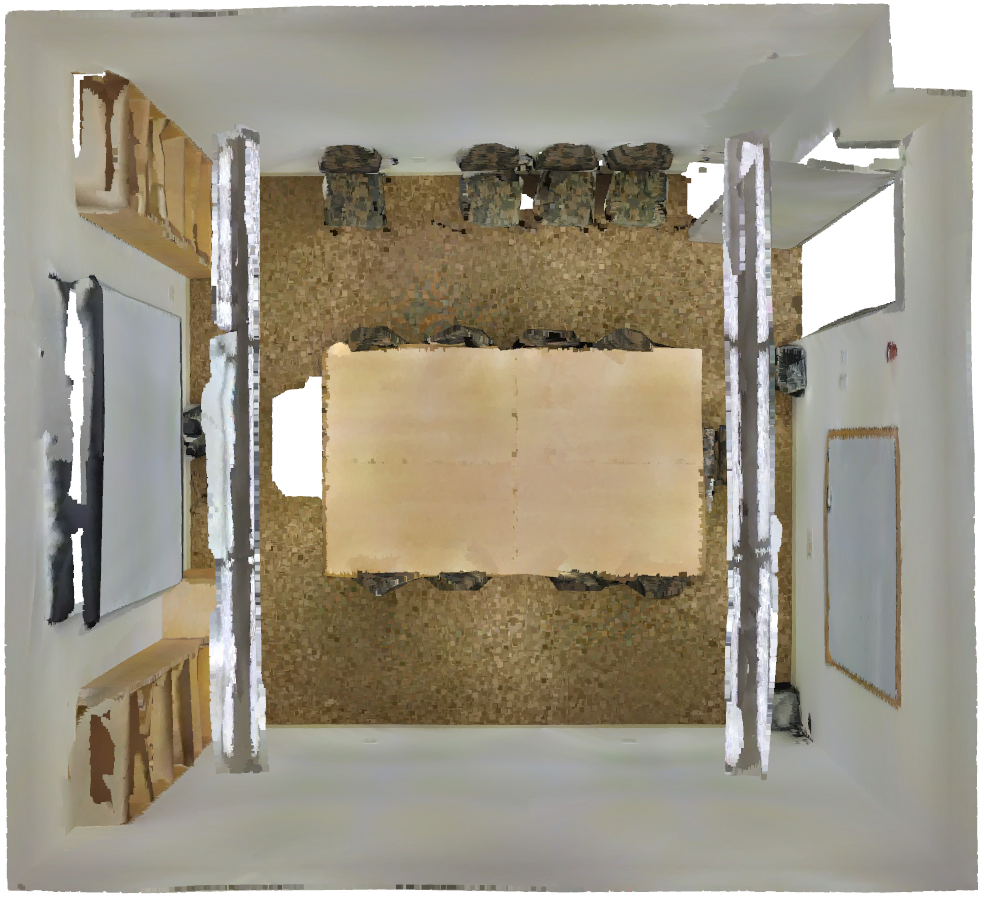}}\hfil
      \subfloat{ \includegraphics[width=0.22\linewidth,height=1in]{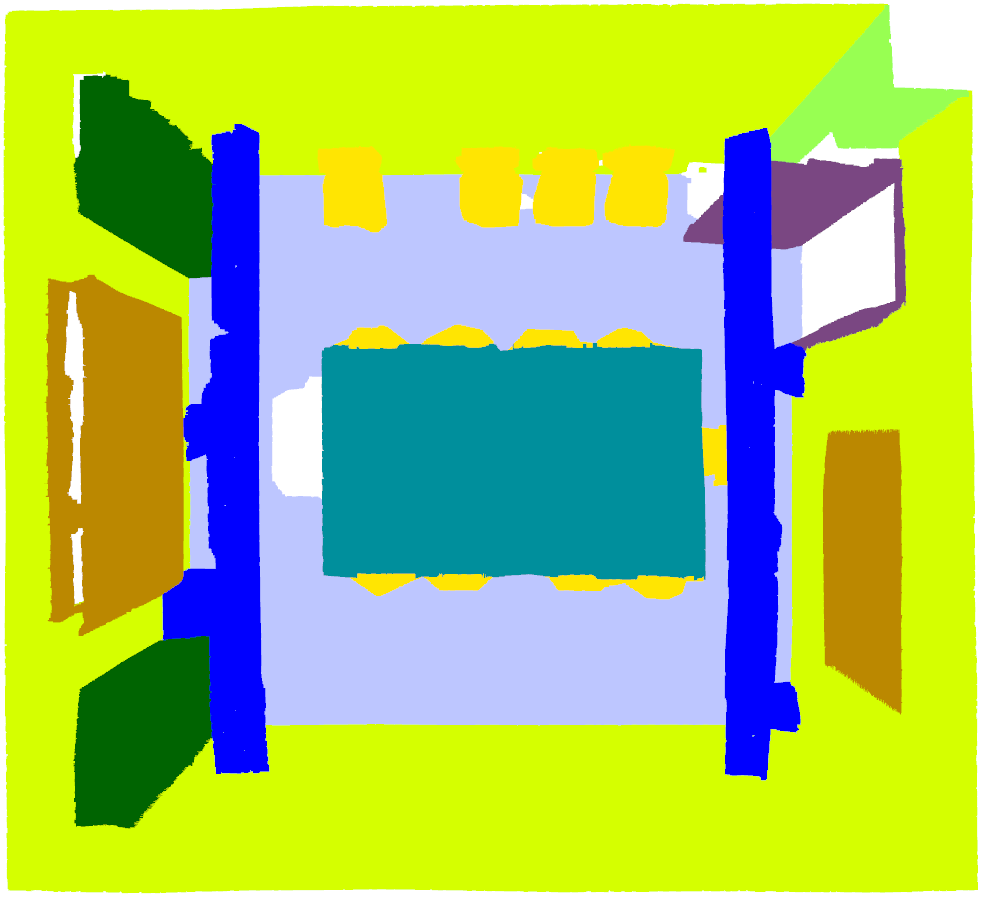}}\hfil
      \subfloat{\includegraphics[width=0.22\linewidth,height=1in]{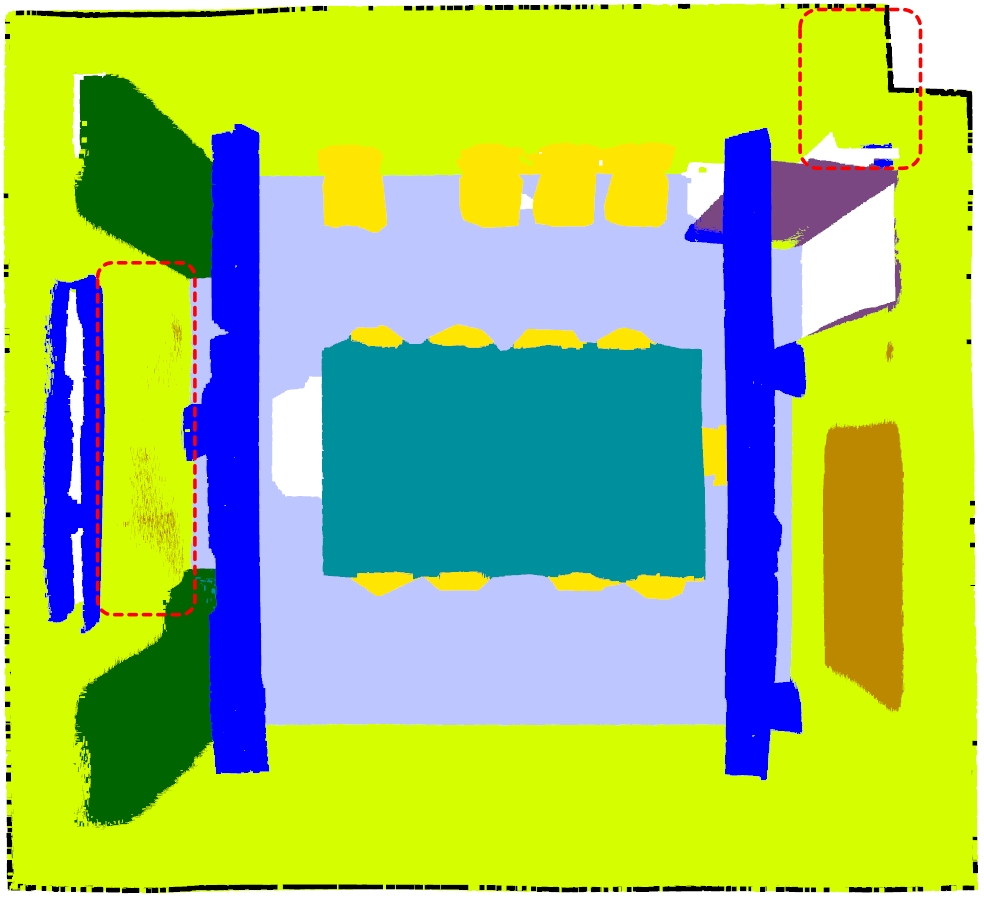}}\hfil
      \subfloat{  \includegraphics[width=0.22\linewidth,height=1in]{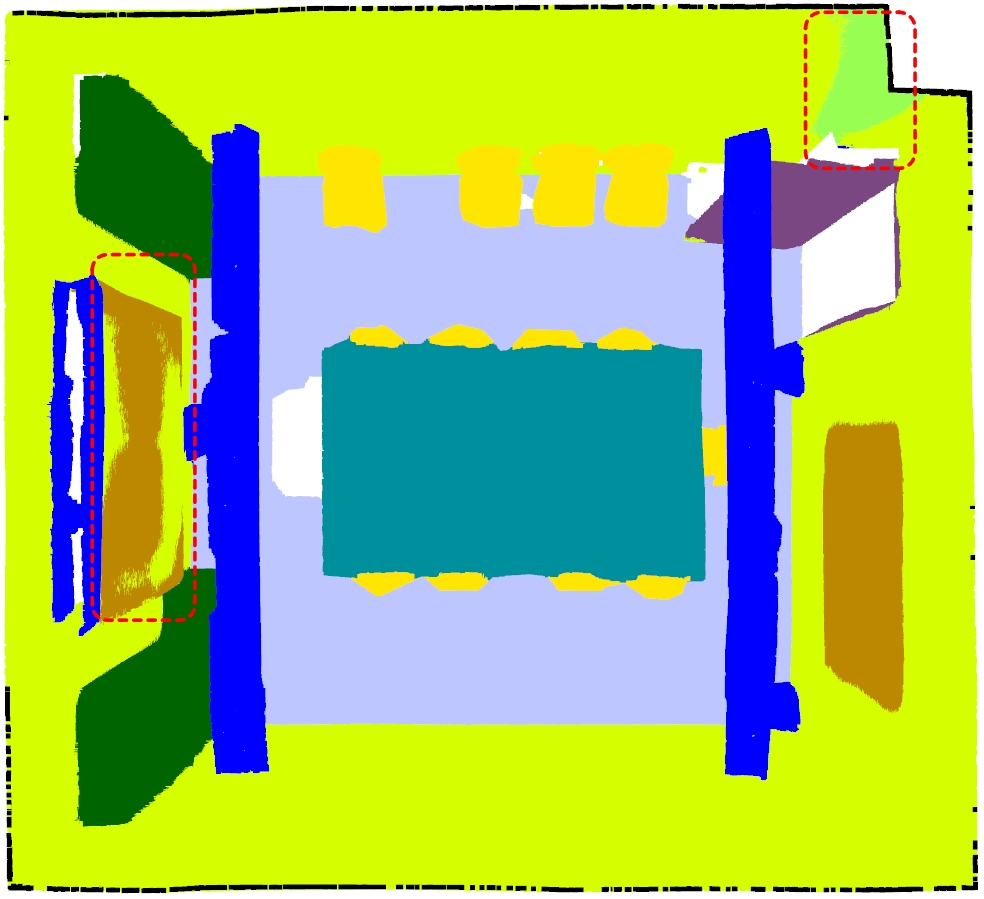}}\hfil
      
      
      \subfloat{\includegraphics[width=0.22\linewidth,height=1in]{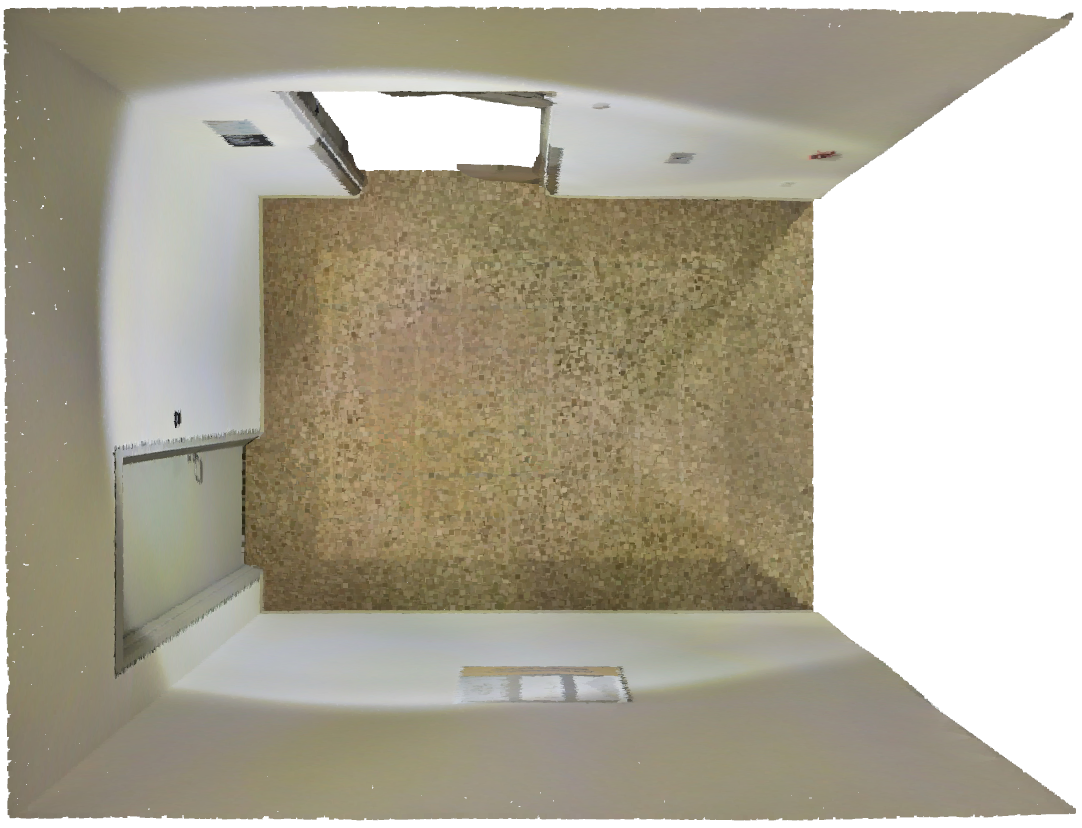}}\hfil
       \subfloat{ \includegraphics[width=0.22\linewidth,height=1in]{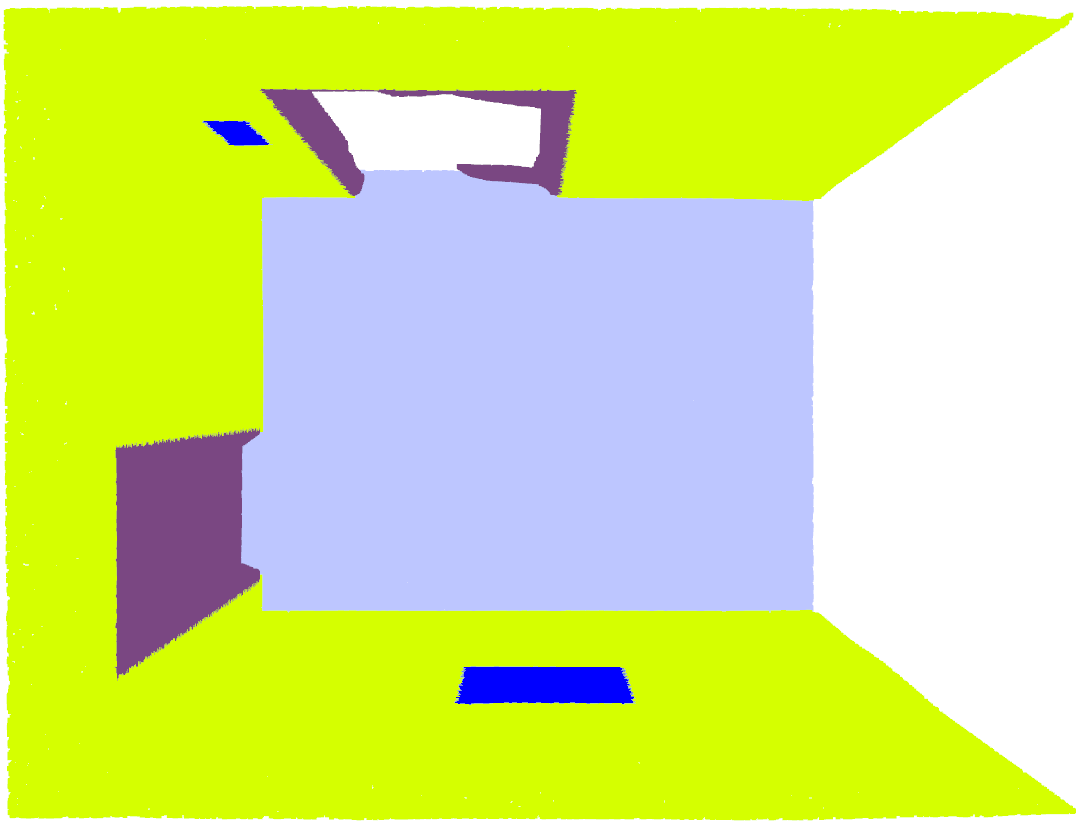}}\hfil
      \subfloat{\includegraphics[width=0.22\linewidth,height=1in]{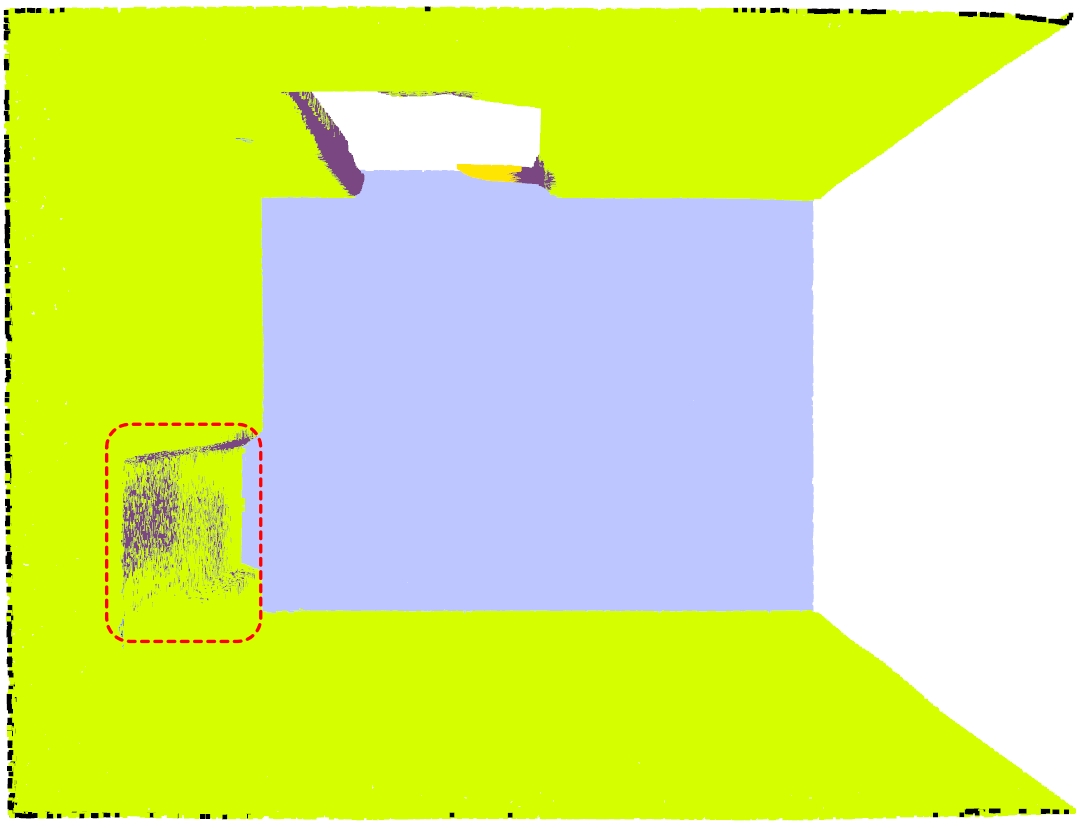}}\hfil
      \subfloat{ \includegraphics[width=0.22\linewidth,height=1in]{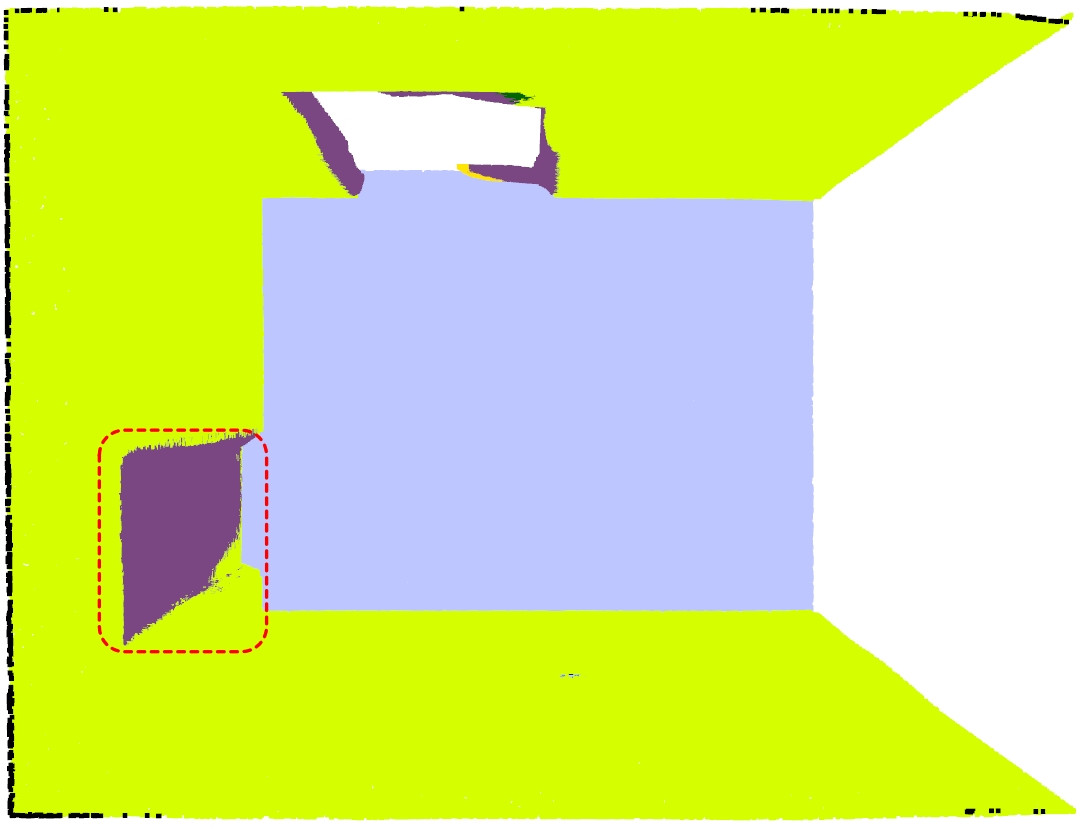}}\hfil

      \subfloat{\includegraphics[width=0.22\linewidth,height=1in]{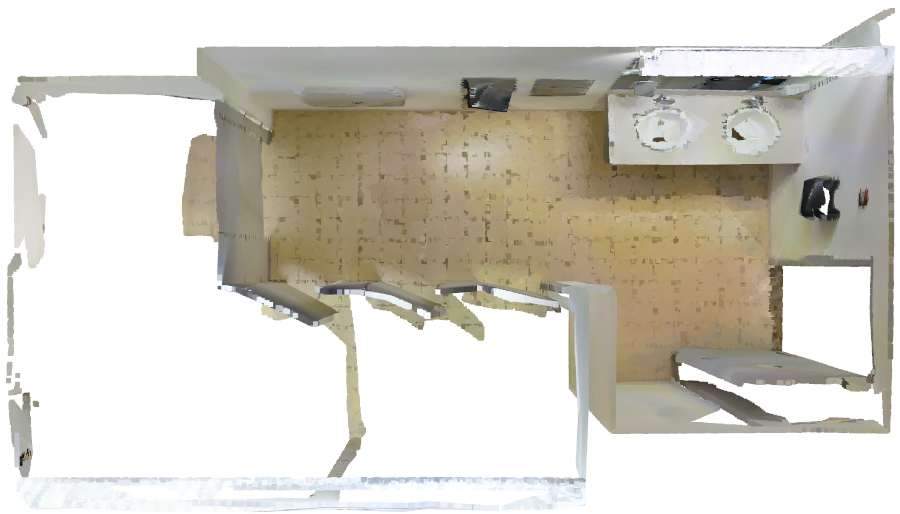}}\hfil
   \subfloat{ \includegraphics[width=0.22\linewidth,height=1in]{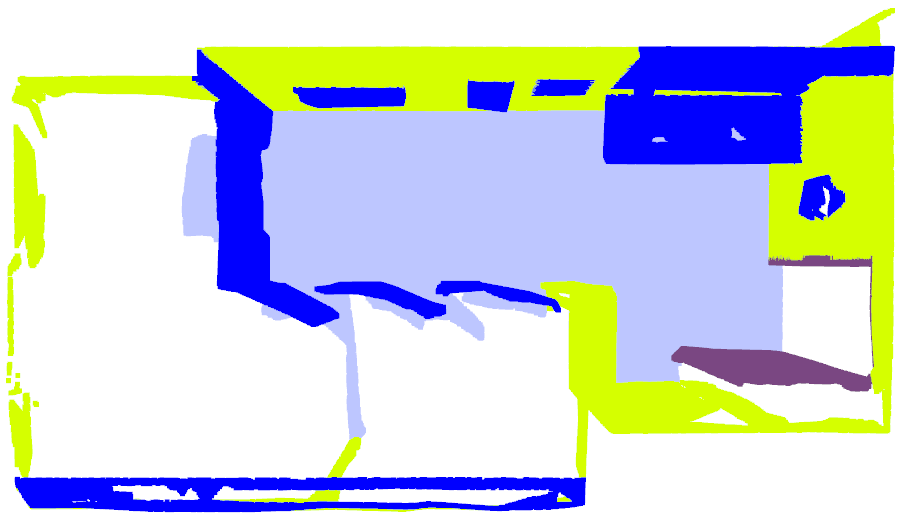}}\hfil
     \subfloat{\includegraphics[width=0.22\linewidth,height=1in]{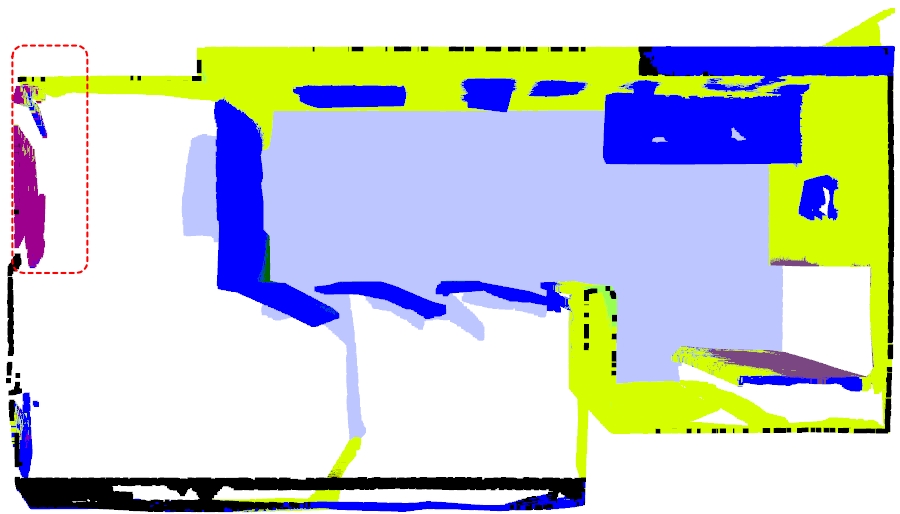}}\hfil
   \subfloat{      \includegraphics[width=0.22\linewidth,height=1in]{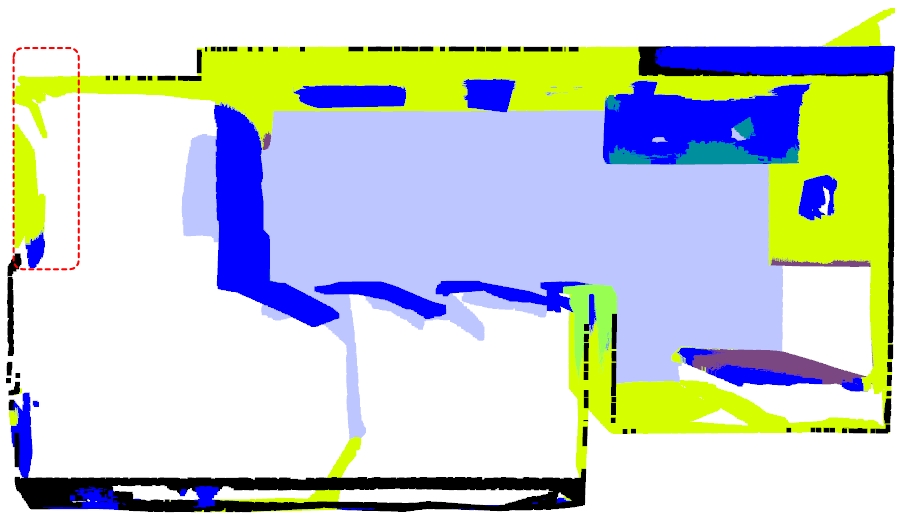}}\hfil
      
  \caption{Segmentation examples on S3DIS~\cite{armeni20163d}. We highlight the region where our method produces better predictions than the Baseline by the red box. Best viewed in color. Left to Right: Room, Ground Truth, Baseline, Ours.}
  \label{a:fig:s3dis} 
\end{figure*}

In addition, we provide 
some visualization comparisons on S3DIS~\cite{armeni20163d} in Figure~\ref{a:fig:s3dis}. For S3DIS, we compare our method and the baseline model (\textit{i.e.,} no edge-to-edge interaction used). We can observe that our model generates more accurate prediction results for some objects (marked with {red} dotted bounding box), like {door} (the $3^{nd}$ and $5^{th}$ examples), {board} ($4^{th}$), {bookcase} ($1^{st}$), wall ( $6^{th}$), and column ($2^{nd}$ and $4^{th}$). The comparisons on S3DIS also support the analysis above.

\section{Conclusion}
In this paper, we propose a novel adaptive edge-to-edge interaction learning scheme for point cloud analysis. In detail, to achieve it, we first develop a new module, \myie{\module}, and then to model the local structure more thoroughly, we further extend the \module\ to a symmetric version, \myie{\smodule}. Through learning the interaction between edges, the module makes the point-to-point relation aware of the local shape, which is beneficial to capture the discriminative local structure information.  To examine the effectiveness of the proposed module, we design two networks, \myie{\netcls\ and \netseg}, for point cloud classification and segmentation, respectively. The experimental results on several typical point cloud tasks and ablations show the models' capability of representing the local structure.
\bibliographystyle{ieee_fullname}
\bibliography{egbib}

\begin{thebibliography}{10}\itemsep=-1pt

\bibitem{armeni20163d}
Iro Armeni, Ozan Sener, Amir~R Zamir, Helen Jiang, Ioannis Brilakis, Martin
  Fischer, and Silvio Savarese.
\newblock 3d semantic parsing of large-scale indoor spaces.
\newblock In {\em Proceedings of the IEEE Conference on Computer Vision and
  Pattern Recognition}, pages 1534--1543, 2016.

\bibitem{atzmon2018point}
Matan Atzmon, Haggai Maron, and Yaron Lipman.
\newblock Point convolutional neural networks by extension operators.
\newblock {\em ACM Trans. Graph.}, 37(4), 2018.

\bibitem{semantickitti}
Jens Behley, Martin Garbade, Andres Milioto, Jan Quenzel, Sven Behnke, Cyrill
  Stachniss, and Jurgen Gall.
\newblock Semantickitti: A dataset for semantic scene understanding of lidar
  sequences.
\newblock In {\em Proceedings of the IEEE/CVF International Conference on
  Computer Vision}, pages 9297--9307, 2019.

\bibitem{Chen_2022_CVPR}
Jiajing Chen, Burak Kakillioglu, Huantao Ren, and Senem Velipasalar.
\newblock Why discard if you can recycle?: A recycling max pooling module for
  3d point cloud analysis.
\newblock In {\em Proceedings of the IEEE/CVF Conference on Computer Vision and
  Pattern Recognition (CVPR)}, pages 559--567, June 2022.

\bibitem{choy20194d}
Christopher Choy, JunYoung Gwak, and Silvio Savarese.
\newblock 4d spatio-temporal convnets: Minkowski convolutional neural networks.
\newblock In {\em Proceedings of the IEEE Conference on Computer Vision and
  Pattern Recognition}, pages 3075--3084, 2019.

\bibitem{dai2017scannet}
Angela Dai, Angel~X Chang, Manolis Savva, Maciej Halber, Thomas Funkhouser, and
  Matthias Nie{\ss}ner.
\newblock Scannet: Richly-annotated 3d reconstructions of indoor scenes.
\newblock In {\em Proceedings of the IEEE Conference on Computer Vision and
  Pattern Recognition}, pages 5828--5839, 2017.

\bibitem{9740525}
Hehe Fan, Yi Yang, and Mohan Kankanhalli.
\newblock Point spatio-temporal transformer networks for point cloud video
  modeling.
\newblock {\em IEEE Transactions on Pattern Analysis and Machine Intelligence},
  pages 1--1, 2022.

\bibitem{8821313}
Shang-Hua Gao, Ming-Ming Cheng, Kai Zhao, Xin-Yu Zhang, Ming-Hsuan Yang, and
  Philip Torr.
\newblock Res2net: A new multi-scale backbone architecture.
\newblock {\em IEEE Transactions on Pattern Analysis and Machine Intelligence},
  43(2):652--662, 2021.

\bibitem{graham20183d}
Benjamin Graham, Martin Engelcke, and Laurens Van Der~Maaten.
\newblock 3d semantic segmentation with submanifold sparse convolutional
  networks.
\newblock In {\em Proceedings of the IEEE conference on computer vision and
  pattern recognition}, pages 9224--9232, 2018.

\bibitem{guo2020deep}
Yulan Guo, Hanyun Wang, Qingyong Hu, Hao Liu, Li Liu, and Mohammed Bennamoun.
\newblock Deep learning for 3d point clouds: A survey.
\newblock {\em IEEE transactions on pattern analysis and machine intelligence},
  2020.

\bibitem{han2020point2node}
Wenkai Han, Chenglu Wen, Cheng Wang, Xin Li, and Qing Li.
\newblock Point2node: Correlation learning of dynamic-node for point cloud
  feature modeling.
\newblock In {\em Proceedings of the AAAI Conference on Artificial
  Intelligence}, volume~34, pages 10925--10932, 2020.

\bibitem{he2016deep}
Kaiming He, Xiangyu Zhang, Shaoqing Ren, and Jian Sun.
\newblock Deep residual learning for image recognition.
\newblock In {\em Proceedings of the IEEE conference on computer vision and
  pattern recognition}, pages 770--778, 2016.

\bibitem{hu2020randla}
Qingyong Hu, Bo Yang, Linhai Xie, Stefano Rosa, Yulan Guo, Zhihua Wang, Niki
  Trigoni, and Andrew Markham.
\newblock Randla-net: Efficient semantic segmentation of large-scale point
  clouds.
\newblock In {\em Proceedings of the IEEE/CVF Conference on Computer Vision and
  Pattern Recognition}, pages 11108--11117, 2020.

\bibitem{9440696}
Qingyong Hu, Bo Yang, Linhai Xie, Stefano Rosa, Yulan Guo, Zhihua Wang, Niki
  Trigoni, and Andrew Markham.
\newblock Learning semantic segmentation of large-scale point clouds with
  random sampling.
\newblock {\em IEEE Transactions on Pattern Analysis and Machine Intelligence},
  pages 1--1, 2021.

\bibitem{hu2021bpnet}
Wenbo Hu, Hengshuang Zhao, Li Jiang, Jiaya Jia, and Tien-Tsin Wong.
\newblock Bidirectional projection network for cross dimension scene
  understanding.
\newblock In {\em CVPR}, 2021.

\bibitem{hu2021vmnet}
Zeyu Hu, Xuyang Bai, Jiaxiang Shang, Runze Zhang, Jiayu Dong, Xin Wang,
  Guangyuan Sun, Hongbo Fu, and Chiew-Lan Tai.
\newblock Vmnet: Voxel-mesh network for geodesic-aware 3d semantic
  segmentation.
\newblock In {\em Proceedings of the IEEE/CVF International Conference on
  Computer Vision}, pages 15488--15498, 2021.

\bibitem{hu2020jsenet}
Zeyu Hu, Mingmin Zhen, Xuyang Bai, Hongbo Fu, and Chiew-lan Tai.
\newblock Jsenet: Joint semantic segmentation and edge detection network for 3d
  point clouds.
\newblock In {\em ECCV}, pages 222--239, 2020.

\bibitem{hua2018pointwise}
Binh-Son Hua, Minh-Khoi Tran, and Sai-Kit Yeung.
\newblock Pointwise convolutional neural networks.
\newblock In {\em Proceedings of the IEEE Conference on Computer Vision and
  Pattern Recognition}, pages 984--993, 2018.

\bibitem{jiang2019hierarchical}
Li Jiang, Hengshuang Zhao, Shu Liu, Xiaoyong Shen, Chi-Wing Fu, and Jiaya Jia.
\newblock Hierarchical point-edge interaction network for point cloud semantic
  segmentation.
\newblock In {\em Proceedings of the IEEE International Conference on Computer
  Vision}, pages 10433--10441, 2019.

\bibitem{kim2014semantic}
David~Inkyu Kim and Gaurav~S Sukhatme.
\newblock Semantic labeling of 3d point clouds with object affordance for robot
  manipulation.
\newblock In {\em 2014 IEEE International Conference on Robotics and Automation
  (ICRA)}, pages 5578--5584. IEEE, 2014.

\bibitem{komarichev2019cnn}
Artem Komarichev, Zichun Zhong, and Jing Hua.
\newblock A-cnn: Annularly convolutional neural networks on point clouds.
\newblock In {\em Proceedings of the IEEE Conference on Computer Vision and
  Pattern Recognition}, pages 7421--7430, 2019.

\bibitem{kundu2020virtual}
Abhijit Kundu, Xiaoqi Yin, Alireza Fathi, David Ross, Brian Brewington, Thomas
  Funkhouser, and Caroline Pantofaru.
\newblock Virtual multi-view fusion for 3d semantic segmentation.
\newblock In {\em European Conference on Computer Vision}, pages 518--535.
  Springer, 2020.

\bibitem{Lai_2022_CVPR}
Xin Lai, Jianhui Liu, Li Jiang, Liwei Wang, Hengshuang Zhao, Shu Liu, Xiaojuan
  Qi, and Jiaya Jia.
\newblock Stratified transformer for 3d point cloud segmentation.
\newblock In {\em Proceedings of the IEEE/CVF Conference on Computer Vision and
  Pattern Recognition (CVPR)}, pages 8500--8509, June 2022.

\bibitem{landrieu2018large}
Loic Landrieu and Martin Simonovsky.
\newblock Large-scale point cloud semantic segmentation with superpoint graphs.
\newblock In {\em Proceedings of the IEEE Conference on Computer Vision and
  Pattern Recognition}, pages 4558--4567, 2018.

\bibitem{lang2020samplenet}
Itai Lang, Asaf Manor, and Shai Avidan.
\newblock Samplenet: Differentiable point cloud sampling.
\newblock In {\em Proceedings of the IEEE/CVF Conference on Computer Vision and
  Pattern Recognition}, pages 7578--7588, 2020.

\bibitem{Le_2020_CVPR}
Eric-Tuan Le, Iasonas Kokkinos, and Niloy~J. Mitra.
\newblock Going deeper with lean point networks.
\newblock In {\em IEEE/CVF Conference on Computer Vision and Pattern
  Recognition (CVPR)}, June 2020.

\bibitem{Lei_2020_CVPR}
Huan Lei, Naveed Akhtar, and Ajmal Mian.
\newblock Seggcn: Efficient 3d point cloud segmentation with fuzzy spherical
  kernel.
\newblock In {\em Proceedings of the IEEE/CVF Conference on Computer Vision and
  Pattern Recognition}, pages 11611--11620, 2020.

\bibitem{9051667}
Huan Lei, Naveed Akhtar, and Ajmal Mian.
\newblock Spherical kernel for efficient graph convolution on 3d point clouds.
\newblock {\em IEEE Transactions on Pattern Analysis and Machine Intelligence},
  43(10):3664--3680, 2021.

\bibitem{li2019net}
Qing Li, Shaoyang Chen, Cheng Wang, Xin Li, Chenglu Wen, Ming Cheng, and
  Jonathan Li.
\newblock Lo-net: Deep real-time lidar odometry.
\newblock In {\em Proceedings of the IEEE Conference on Computer Vision and
  Pattern Recognition}, pages 8473--8482, 2019.

\bibitem{li2018pointcnn}
Yangyan Li, Rui Bu, Mingchao Sun, Wei Wu, Xinhan Di, and Baoquan Chen.
\newblock Pointcnn: Convolution on x-transformed points.
\newblock In {\em Advances in neural information processing systems}, pages
  820--830, 2018.

\bibitem{lin2020fpconv}
Yiqun Lin, Zizheng Yan, Haibin Huang, Dong Du, Ligang Liu, Shuguang Cui, and
  Xiaoguang Han.
\newblock Fpconv: Learning local flattening for point convolution.
\newblock In {\em Proceedings of the IEEE/CVF Conference on Computer Vision and
  Pattern Recognition}, pages 4293--4302, 2020.

\bibitem{3dgcn}
Zhi-Hao Lin, Sheng-Yu Huang, and Yu-Chiang~Frank Wang.
\newblock Learning of 3d graph convolution networks for point cloud analysis.
\newblock {\em IEEE Transactions on Pattern Analysis and Machine Intelligence},
  44(8):4212--4224, 2022.

\bibitem{liu2019dynamic}
Jinxian Liu, Bingbing Ni, Caiyuan Li, Jiancheng Yang, and Qi Tian.
\newblock Dynamic points agglomeration for hierarchical point sets learning.
\newblock In {\em Proceedings of the IEEE International Conference on Computer
  Vision}, pages 7546--7555, 2019.

\bibitem{liu2019point2sequence}
Xinhai Liu, Zhizhong Han, Yu-Shen Liu, and Matthias Zwicker.
\newblock Point2sequence: Learning the shape representation of 3d point clouds
  with an attention-based sequence to sequence network.
\newblock In {\em Proceedings of the AAAI Conference on Artificial
  Intelligence}, volume~33, pages 8778--8785, 2019.

\bibitem{liu2019densepoint}
Yongcheng Liu, Bin Fan, Gaofeng Meng, Jiwen Lu, Shiming Xiang, and Chunhong
  Pan.
\newblock Densepoint: Learning densely contextual representation for efficient
  point cloud processing.
\newblock In {\em Proceedings of the IEEE International Conference on Computer
  Vision}, pages 5239--5248, 2019.

\bibitem{liu2019relation}
Yongcheng Liu, Bin Fan, Shiming Xiang, and Chunhong Pan.
\newblock Relation-shape convolutional neural network for point cloud analysis.
\newblock In {\em Proceedings of the IEEE Conference on Computer Vision and
  Pattern Recognition}, pages 8895--8904, 2019.

\bibitem{liu2020closerlook3d}
Ze Liu, Han Hu, Yue Cao, Zheng Zhang, and Xin Tong.
\newblock A closer look at local aggregation operators in point cloud analysis.
\newblock {\em ECCV}, 2020.

\bibitem{loshchilov2016sgdr}
Ilya Loshchilov and Frank Hutter.
\newblock Sgdr: Stochastic gradient descent with warm restarts.
\newblock {\em arXiv preprint arXiv:1608.03983}, 2016.

\bibitem{maturana2015voxnet}
Daniel Maturana and Sebastian Scherer.
\newblock Voxnet: A 3d convolutional neural network for real-time object
  recognition.
\newblock In {\em 2015 IEEE/RSJ International Conference on Intelligent Robots
  and Systems (IROS)}, pages 922--928. IEEE, 2015.

\bibitem{9356353}
Shervin Minaee, Yuri Boykov, Fatih Porikli, Antonio Plaza, Nasser Kehtarnavaz,
  and Demetri Terzopoulos.
\newblock Image segmentation using deep learning: A survey.
\newblock {\em IEEE Transactions on Pattern Analysis and Machine Intelligence},
  44(7):3523--3542, 2022.

\bibitem{Nezhadarya_2020_CVPR}
Ehsan Nezhadarya, Ehsan Taghavi, Ryan Razani, Bingbing Liu, and Jun Luo.
\newblock Adaptive hierarchical down-sampling for point cloud classification.
\newblock In {\em IEEE/CVF Conference on Computer Vision and Pattern
  Recognition (CVPR)}, June 2020.

\bibitem{Nie_2022_CVPR}
Dong Nie, Rui Lan, Ling Wang, and Xiaofeng Ren.
\newblock Pyramid architecture for multi-scale processing in point cloud
  segmentation.
\newblock In {\em Proceedings of the IEEE/CVF Conference on Computer Vision and
  Pattern Recognition (CVPR)}, pages 17284--17294, June 2022.

\bibitem{nie2021differentiable}
Xing Nie, Yongcheng Liu, Shaohong Chen, Jianlong Chang, Chunlei Huo, Gaofeng
  Meng, Qi Tian, Weiming Hu, and Chunhong Pan.
\newblock Differentiable convolution search for point cloud processing.
\newblock In {\em Proceedings of the IEEE/CVF International Conference on
  Computer Vision}, pages 7437--7446, 2021.

\bibitem{fpt}
Chunghyun Park, Yoonwoo Jeong, Minsu Cho, and Jaesik Park.
\newblock Fast point transformer.
\newblock In {\em Proceedings of the {IEEE/CVF} Conference on Computer Vision
  and Pattern Recognition (CVPR)}, pages 16949--16958, June 2022.

\bibitem{qi2017pointnet}
Charles~R Qi, Hao Su, Kaichun Mo, and Leonidas~J Guibas.
\newblock Pointnet: Deep learning on point sets for 3d classification and
  segmentation.
\newblock In {\em Proceedings of the IEEE conference on computer vision and
  pattern recognition}, pages 652--660, 2017.

\bibitem{qi2017pointnet++}
Charles~Ruizhongtai Qi, Li Yi, Hao Su, and Leonidas~J Guibas.
\newblock Pointnet++: Deep hierarchical feature learning on point sets in a
  metric space.
\newblock In {\em Advances in neural information processing systems}, pages
  5099--5108, 2017.

\bibitem{9661313}
Shi Qiu, Saeed Anwar, and Nick Barnes.
\newblock Pnp-3d: A plug-and-play for 3d point clouds.
\newblock {\em IEEE Transactions on Pattern Analysis and Machine Intelligence},
  pages 1--1, 2021.

\bibitem{ran2021learning}
Haoxi Ran, Wei Zhuo, Jun Liu, and Li Lu.
\newblock Learning inner-group relations on point clouds.
\newblock In {\em Proceedings of the IEEE/CVF International Conference on
  Computer Vision}, pages 15477--15487, 2021.

\bibitem{9018080}
Shaoshuai Shi, Zhe Wang, Jianping Shi, Xiaogang Wang, and Hongsheng Li.
\newblock From points to parts: 3d object detection from point cloud with
  part-aware and part-aggregation network.
\newblock {\em IEEE Transactions on Pattern Analysis and Machine Intelligence},
  43(8):2647--2664, 2021.

\bibitem{tang2020searching}
Haotian Tang, Zhijian Liu, Shengyu Zhao, Yujun Lin, Ji Lin, Hanrui Wang, and
  Song Han.
\newblock Searching efficient 3d architectures with sparse point-voxel
  convolution.
\newblock 2020.

\bibitem{Tang_2022_CVPR}
Liyao Tang, Yibing Zhan, Zhe Chen, Baosheng Yu, and Dacheng Tao.
\newblock Contrastive boundary learning for point cloud segmentation.
\newblock In {\em Proceedings of the IEEE/CVF Conference on Computer Vision and
  Pattern Recognition (CVPR)}, pages 8489--8499, June 2022.

\bibitem{thomas2019kpconv}
Hugues Thomas, Charles~R Qi, Jean-Emmanuel Deschaud, Beatriz Marcotegui,
  Fran{\c{c}}ois Goulette, and Leonidas~J Guibas.
\newblock Kpconv: Flexible and deformable convolution for point clouds.
\newblock In {\em Proceedings of the IEEE International Conference on Computer
  Vision}, pages 6411--6420, 2019.

\bibitem{scanobjectnn}
Mikaela~Angelina Uy, Quang-Hieu Pham, Binh-Son Hua, Duc~Thanh Nguyen, and
  Sai-Kit Yeung.
\newblock Revisiting point cloud classification: A new benchmark dataset and
  classification model on real-world data.
\newblock In {\em International Conference on Computer Vision (ICCV)}, 2019.

\bibitem{vaswani2017attention}
Ashish Vaswani, Noam Shazeer, Niki Parmar, Jakob Uszkoreit, Llion Jones,
  Aidan~N. Gomez, \L{}ukasz Kaiser, and Illia Polosukhin.
\newblock Attention is all you need.
\newblock In {\em Advances in Neural Information Processing Systems}, page
  6000–6010, 2017.

\bibitem{trans3d}
Jiayun Wang, Rudrasis Chakraborty, and Stella~X. Yu.
\newblock Transformer for 3d point clouds.
\newblock {\em IEEE Transactions on Pattern Analysis and Machine Intelligence},
  44(8):4419--4431, 2022.

\bibitem{wang2019graph}
Lei Wang, Yuchun Huang, Yaolin Hou, Shenman Zhang, and Jie Shan.
\newblock Graph attention convolution for point cloud semantic segmentation.
\newblock In {\em Proceedings of the IEEE Conference on Computer Vision and
  Pattern Recognition}, pages 10296--10305, 2019.

\bibitem{wang2018deep}
Shenlong Wang, Simon Suo, Wei-Chiu Ma, Andrei Pokrovsky, and Raquel Urtasun.
\newblock Deep parametric continuous convolutional neural networks.
\newblock In {\em Proceedings of the IEEE Conference on Computer Vision and
  Pattern Recognition}, pages 2589--2597, 2018.

\bibitem{9320524}
Wenguan Wang, Qiuxia Lai, Huazhu Fu, Jianbing Shen, Haibin Ling, and Ruigang
  Yang.
\newblock Salient object detection in the deep learning era: An in-depth
  survey.
\newblock {\em IEEE Transactions on Pattern Analysis and Machine Intelligence},
  44(6):3239--3259, 2022.

\bibitem{wang2019exploiting}
Xu Wang, Jingming He, and Lin Ma.
\newblock Exploiting local and global structure for point cloud semantic
  segmentation with contextual point representations.
\newblock In {\em Advances in Neural Information Processing Systems}, pages
  4571--4581, 2019.

\bibitem{wang2019dynamic}
Yue Wang, Yongbin Sun, Ziwei Liu, Sanjay~E Sarma, Michael~M Bronstein, and
  Justin~M Solomon.
\newblock Dynamic graph cnn for learning on point clouds.
\newblock {\em Acm Transactions On Graphics (tog)}, 38(5):1--12, 2019.

\bibitem{9735342}
Xin Wen, Peng Xiang, Zhizhong Han, Yan-Pei Cao, Pengfei Wan, Wen Zheng, and
  Yu-Shen Liu.
\newblock Pmp-net++: Point cloud completion by transformer-enhanced multi-step
  point moving paths.
\newblock {\em IEEE Transactions on Pattern Analysis and Machine Intelligence},
  pages 1--1, 2022.

\bibitem{wongefficient}
Chi-Chong Wong and Chi-Man Vong.
\newblock Efficient outdoor 3d point cloud semantic segmentation for critical
  road objects and distributed contexts.
\newblock In {\em European Conference on Computer Vision}, pages 499--514,
  2020.

\bibitem{wu2019pointconv}
Wenxuan Wu, Zhongang Qi, and Li Fuxin.
\newblock Pointconv: Deep convolutional networks on 3d point clouds.
\newblock In {\em Proceedings of the IEEE Conference on Computer Vision and
  Pattern Recognition}, pages 9621--9630, 2019.

\bibitem{wu20153d}
Zhirong Wu, Shuran Song, Aditya Khosla, Fisher Yu, Linguang Zhang, Xiaoou Tang,
  and Jianxiong Xiao.
\newblock 3d shapenets: A deep representation for volumetric shapes.
\newblock In {\em Proceedings of the IEEE conference on computer vision and
  pattern recognition}, pages 1912--1920, 2015.

\bibitem{zhang2021BowPooling}
Zhouhui~Lian Xiang~Zhang, Xiao~Sun.
\newblock Bow pooling: A plug-and-play unit for feature aggregation of point
  clouds.
\newblock {\em AAAI 2021}, 2021.

\bibitem{paconv}
Mutian Xu, Runyu Ding, Hengshuang Zhao, and Xiaojuan Qi.
\newblock Paconv: Position adaptive convolution with dynamic kernel assembling
  on point clouds.
\newblock In {\em Proceedings of the IEEE/CVF Conference on Computer Vision and
  Pattern Recognition (CVPR)}, pages 3173--3182, June 2021.

\bibitem{xu2020grid}
Qiangeng Xu, Xudong Sun, Cho-Ying Wu, Panqu Wang, and Ulrich Neumann.
\newblock Grid-gcn for fast and scalable point cloud learning.
\newblock In {\em Proceedings of the IEEE/CVF Conference on Computer Vision and
  Pattern Recognition}, pages 5661--5670, 2020.

\bibitem{xu2018spidercnn}
Yifan Xu, Tianqi Fan, Mingye Xu, Long Zeng, and Yu Qiao.
\newblock Spidercnn: Deep learning on point sets with parameterized
  convolutional filters.
\newblock In {\em Proceedings of the European Conference on Computer Vision
  (ECCV)}, pages 87--102, 2018.

\bibitem{yan2020pointasnl}
Xu Yan, Chaoda Zheng, Zhen Li, Sheng Wang, and Shuguang Cui.
\newblock Pointasnl: Robust point clouds processing using nonlocal neural
  networks with adaptive sampling.
\newblock In {\em Proceedings of the IEEE/CVF Conference on Computer Vision and
  Pattern Recognition}, pages 5589--5598, 2020.

\bibitem{yang2019modeling}
Jiancheng Yang, Qiang Zhang, Bingbing Ni, Linguo Li, Jinxian Liu, Mengdie Zhou,
  and Qi Tian.
\newblock Modeling point clouds with self-attention and gumbel subset sampling.
\newblock In {\em Proceedings of the IEEE Conference on Computer Vision and
  Pattern Recognition}, pages 3323--3332, 2019.

\bibitem{Yi16}
Li Yi, Vladimir~G. Kim, Duygu Ceylan, I-Chao Shen, Mengyan Yan, Hao Su, Cewu
  Lu, Qixing Huang, Alla Sheffer, and Leonidas Guibas.
\newblock A scalable active framework for region annotation in 3d shape
  collections.
\newblock {\em SIGGRAPH Asia}, 2016.

\bibitem{zhang2019shellnet}
Zhiyuan Zhang, Binh-Son Hua, and Sai-Kit Yeung.
\newblock Shellnet: Efficient point cloud convolutional neural networks using
  concentric shells statistics.
\newblock In {\em Proceedings of the IEEE International Conference on Computer
  Vision}, pages 1607--1616, 2019.

\bibitem{zhao2019pointweb}
Hengshuang Zhao, Li Jiang, Chi-Wing Fu, and Jiaya Jia.
\newblock Pointweb: Enhancing local neighborhood features for point cloud
  processing.
\newblock In {\em Proceedings of the IEEE Conference on Computer Vision and
  Pattern Recognition}, pages 5565--5573, 2019.

\bibitem{ptransformer}
Hengshuang Zhao, Li Jiang, Jiaya Jia, Philip~H.S. Torr, and Vladlen Koltun.
\newblock Point transformer.
\newblock In {\em Proceedings of the IEEE/CVF International Conference on
  Computer Vision (ICCV)}, pages 16259--16268, October 2021.

\bibitem{zhou2021adaptive}
Haoran Zhou, Yidan Feng, Mingsheng Fang, Mingqiang Wei, Jing Qin, and Tong Lu.
\newblock Adaptive graph convolution for point cloud analysis.
\newblock In {\em Proceedings of the IEEE/CVF International Conference on
  Computer Vision}, pages 4965--4974, 2021.

\bibitem{zhu2021cylindrical}
Xinge Zhu, Hui Zhou, Tai Wang, Fangzhou Hong, Yuexin Ma, Wei Li, Hongsheng Li,
  and Dahua Lin.
\newblock Cylindrical and asymmetrical 3d convolution networks for lidar
  segmentation.
\newblock In {\em Proceedings of the IEEE/CVF conference on computer vision and
  pattern recognition}, pages 9939--9948, 2021.

\end{thebibliography}
%



%




\end{document}